\documentclass[10pt,draftcls,onecolumn]{IEEEtran}

\IEEEoverridecommandlockouts                              
\usepackage{graphicx}
\usepackage{epsfig}
\usepackage{array}
\usepackage{paralist}
\usepackage{verbatim}
\usepackage{caption}
\usepackage{subcaption}
\usepackage{epstopdf}
\usepackage{color,soul}
\definecolor{lightblue}{rgb}{.90,.95,1}
\sethlcolor{yellow}

\usepackage{amsfonts,amssymb } 

\usepackage{epsfig} 
\usepackage{times} 
\usepackage{amsmath} 
\usepackage{amssymb}  
\usepackage{algorithm}
\usepackage{algpseudocode}
\usepackage{tabularx}
\usepackage{booktabs}
\makeatletter
\renewcommand{\ALG@beginalgorithmic}{\small}
\makeatother

\makeatletter
\renewcommand{\maketag@@@}[1]{\hbox{\m@th\normalsize\normalfont#1}}%
\makeatother

\begin{document}

\title{\huge End-to-End Design \& Validation of a Low-Cost Stewart Platform with Nonlinear Estimation \& Control}


\author{
    Benedictus C. G. Cinun, 
    Tua A. Tamba$^*$, 
    Immanuel R. Santjoko, 
    Xiaofeng Wang, 
    Michael A. Gunarso, 
    Bin Hu 
    \thanks{
    B. C. G. Cinun and B. Hu are with the Department of Electrical and Computer Engineering at University of Houston, Houston, TX 77004, USA. ({\tt\small bgerodacinun@uh.edu})}%
    \thanks{
    T. A. Tamba, I. R. Santjoko, and M. A. Gunarso are with the Department of Electrical Engineering, Faculty of Engineering Technology, Parahyangan Catholic University, Bandung, Indonesia ({\tt\small ttamba@unpar.ac.id, 6152001008@student.unpar.ac.id})}%
    \thanks{
    X. Wang is with Department of Electrical Engineering, University of South Carolina, Columbia, SC 29208, USA ({\tt\small wangxi@cec.sc.edu)}}%
    \thanks{
    B. Hu is with Department of Engineering Technology, Electrical and Computer Engineering, University of Houston, Houston, TX 77004, USA ({\tt\small bhu11@central.uh.edu)}}
}

\maketitle

\begin{abstract}
This paper presents the complete design, control, and experimental validation of a low-cost Stewart platform prototype developed as an affordable yet capable robotic testbed for research and education. The platform combines off-the-shelf components with 3D-printed and custom-fabricated parts to deliver full six-degree-of-freedom (6-DOF) motions using six linear actuators connecting a moving platform to a fixed base. The system software integrates dynamic modeling, data acquisition, and real-time control within a unified framework. A robust trajectory tracking controller based on feedback linearization, augmented with an LQR scheme, compensates for the platform’s nonlinear dynamics to achieve precise motion control. In parallel, an Extended Kalman Filter fuses IMU and actuator encoder feedback to provide accurate and reliable state estimation under sensor noise and external disturbances. Unlike prior efforts that emphasize only isolated aspects such as modeling or control, this work delivers a complete hardware–software platform validated through both simulation and experiments on static and dynamic trajectories. Results demonstrate effective trajectory tracking and real-time state estimation, highlighting the platform’s potential as a cost-effective and versatile tool for advanced research and educational applications.
\end{abstract}



\section{Introduction}
\label{sec:intro}

The Stewart platform is a six-degree-of-freedom (6-DoF) parallel manipulator renowned for its high stiffness, precision, and ability to support large payloads relative to its size~\cite{stewart1965platform,antonov2024,cheng2001advantages}.
These properties make it well-suited for applications requiring precise motion and load-bearing capability, including flight simulators~\cite{wei2022design,huang2016review}, wave compensation in maritime operations~\cite{cai2020model,cai2021sliding,qiu2024modeling}, evaluating the performance of 3D-printed components under simulated ocean wave conditions~\cite{silva2022stewart}, surgical robotics~\cite{kizir2019design,patel2018sprk}, telescope positioning~\cite{li2022decoupling,jikuya2021structure}, and more recently, fine adjustment of segmented mirrors on the James-Webb Space Telescope (JWST)~\cite{liang2022kinematics}.
The breadth of these applications highlights the platform's versatility and ongoing relevance in both industrial and research domains~\cite{patel2012parallel,ghodsian2023mobile,zarebidoki2022review,yang2022review,russo2024review}.

\begin{figure}[!b]
\centering
\subfloat[Acrome]{%
  \includegraphics[width=0.2\textwidth, height = 3cm]{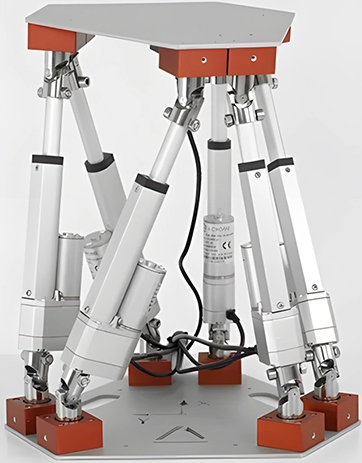}%
  \label{fig:acrome_SP}}
  \hspace{1cm}
\subfloat[PT-Actuators]{%
  \includegraphics[width=0.2\textwidth, height = 3cm]{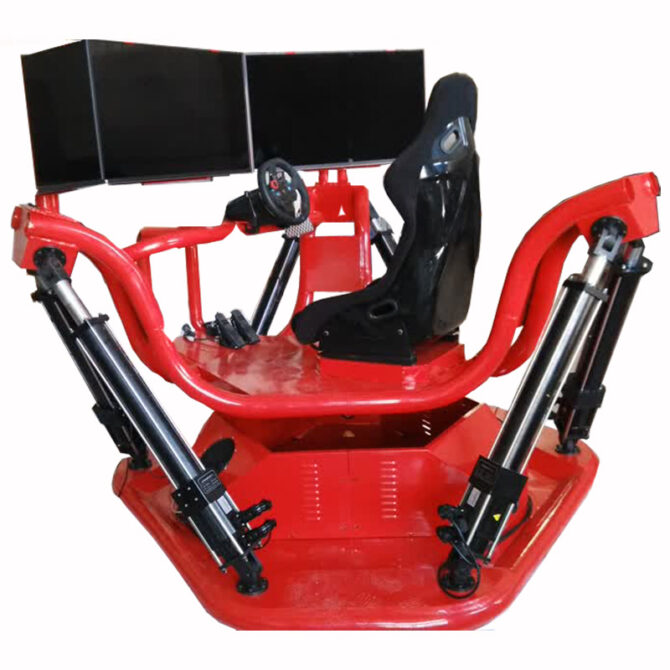}%
  \label{fig:PT_SP}}
  \hspace{1cm}
\subfloat[Motion Systems]{%
  \includegraphics[width=0.2\textwidth, height = 3cm]{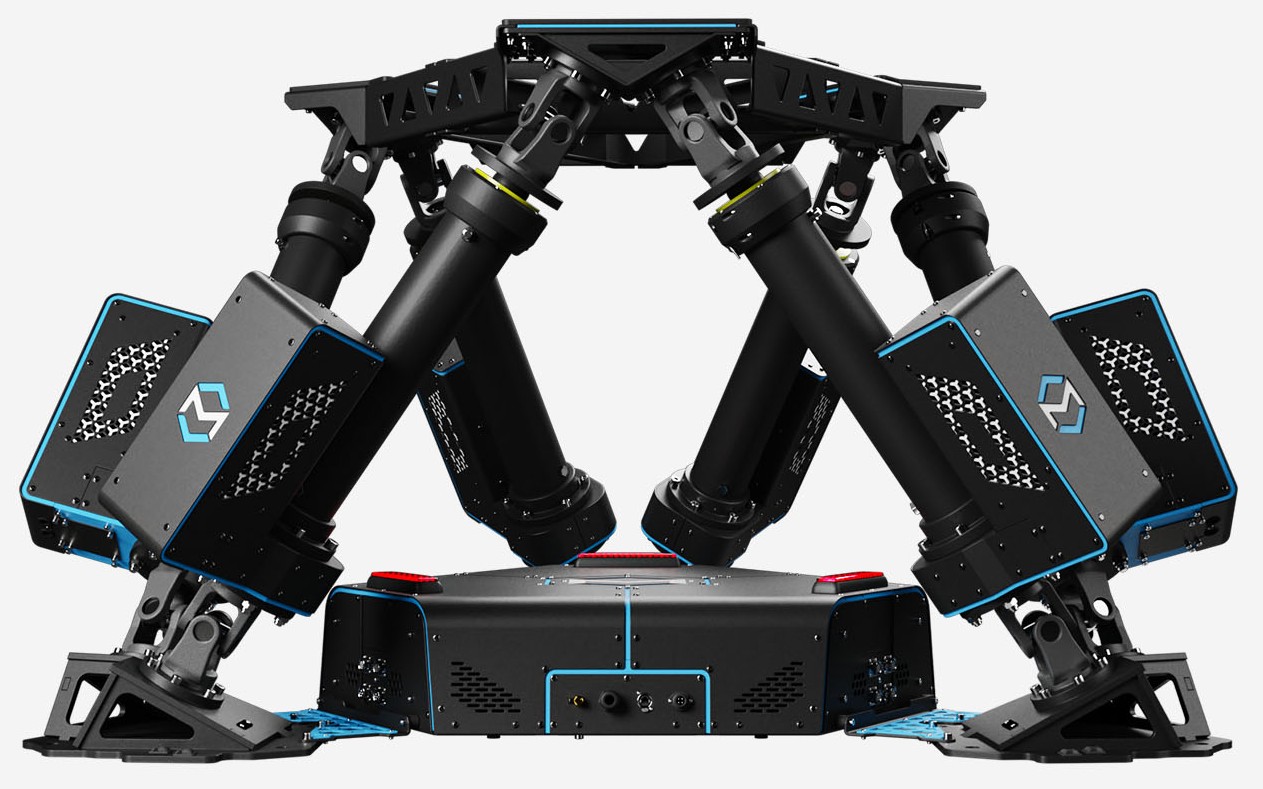}%
  \label{fig:MotionSystems_SP}
}%
\vspace{-3pt}
\caption{Examples of commercially-developed Stewart platforms (sources:~\cite{acromeStewart,ptactuator2025,motionsystems2025}).}
\end{figure}

Access to Stewart platforms for experimental research and education purposes is often limited by high costs, as commercial systems typically use precision actuators, high-resolution sensors, and advanced controllers, placing them beyond the reach of many academic or small-scale laboratories.
For example, the Acrome Stewart platform in Fig.~\ref{fig:acrome_SP} costs around USD $\$8,899$~\cite{acromeStewart}. 
Larger systems for flight simulation or heavy-duty testing purposes are even more expensive, such as the PT-Actuator model in Fig.~\ref{fig:PT_SP} whose cost starts at USD $\$12,000$~\cite{ptactuator2025}, and the Motion Systems platform in Fig.~\ref{fig:MotionSystems_SP} whose cost reaches up to EUR 50,000~\cite{motionsystems2025}.
To address this gap, several low-cost prototypes have been developed using alternative actuation and sensing strategies, including an educational proof-of-concept platform~\cite{venkat2021design}, a surgical motion simulator that replaced linear actuators with servo motors to reduce costs to USD $\$250$~\cite{patel2018sprk}, and a haptic controller driven by hobby servos~\cite{kelesbekov2024stuet}.
However, hobby servos often have limited performance due to low-cost potentiometers and restricted torque.
A cost-conscious Stewart platform was proposed in~\cite{grogan2020low} that uses industrial servomotors for simulator applications to emphasize portability, performance, and the ability to handle human-sized payloads.
In contrast, our design employs linear actuators, which provide higher load capacity and stiffness~\cite{yu2024research}, while integrated potentiometer feedback enables position sensing for closed-loop length control. 
It is worth noting that an addition of elastic springs to the linear actuator has also been proposed, which could act as a low-pass filter to reduce disturbances from actuator backlash and external forces \cite{ophaswongse2018}.
A related low-cost design also used linear actuators~\cite{peterson2020}; however, it relied on an Arduino controller, which limited computational capacity and lacked the design and implementation of advanced control and estimation strategies.
In contrast, our developed Stewart platform incorporates nonlinear control and estimation methods to enable a comprehensive exploration of the functionality of the platform, making it more suitable for both education and advanced experimental research.

For the purpose of implementing advanced control strategies and ensure reliable operation of Stewart platforms, it is essential to develop an accurate system model and state estimation algorithm. 
Early approaches for solving forward kinematics problems in parallel manipulators often relied on closed-form geometric solutions~\cite{liu2018geometric,innocenti1992direct,lin1992closed,parenti2013direct,griffis1989forward,innocenti1996closed} or the Denavit–Hartenberg convention~\cite{shim2022}. 
A modeling approach using Conformal Geometric Algebra method has also been proposed in~\cite{zhu2024conformal} to characterize the relationship between the platform joints and the states of the end-effector.
Although these analytic approaches can provide an exact solution, they are often sensitive to measurement noise and yielded multiple solution configurations.
To resolve the limitation of analytical approach, iterative numerical methods such as Newton–Raphson and Levenberg–Marquardt methods have been proposed in~\cite{bingul2012dynamic, YunfengForwardKine,FKgithub,yang2010forward-Iter1,Merlet-Iter2,jauregui-correa_validation_2015}, but these methods require high-performance machines and accurate initial guesses.
The works in~\cite{Sensor2,ding_stiffness_2014} proposed the use of additional external sensors to improve the accuracy of the model and the state estimation, respectively, to provide simplified kinematics and enhanced robustness of the model at the expense of additional hardware complexity.
Other uses of additional sensors such as indoor GPS~\cite{de2020offline} and camera-based systems~\cite{song2015development} have also been explored to improve the state estimation algorithm, although they are often difficult to set up and calibrate.
The use of observer-based techniques, such as the Unscented Kalman Filter~\cite{MILETOVIC2017102} method, has also been proposed to integrate sensor measurements with kinematic models to reduce measurement noise and improve the reliability of the estimation results. 
Finally, machine learning methods, such as artificial~\cite{parikh2009solvingML2} and feed-forward~\cite{chauhan2022forward} neural networks, have recently been investigated to solve the forward kinematics and pose estimation problems of the Stewart platform.

In addition to modeling and state estimation approaches, various control strategies have been proposed to optimize the Stewart platform under different operational requirements.
These control strategies range from classical PID control~\cite{csumnu2017simulation,bernal2024control,du_nonlinear_2011} to advanced control methods such as model predictive control and adaptive control based on kinematic or dynamic models of the platform~\cite{ahmadi2020nonlinear,NMPC,miunske_model_2018}.
Other designs have also incorporated sliding mode control with feedforward velocity compensation~\cite{cai2021sliding} as well as the combination of optimization-based control~\cite{wang2025linear,pradipta_actuator_2016}.
More recently, data-driven methods based on neural networks~\cite{shi2024real} and reinforcement learning~\cite{yadavari2024addressing} have also been explored to improve the performance of the Stewart platform.
Furthermore, deep Reinforcement Learning has also been used in \cite{yadavari2023} to determine optimal PID gains and validated in a Gazebo simulation environment that is integrated with the Robot Operating System (ROS).
In \cite{chen2023}, a modal space control method was also proposed that incorporates ship motion into the model, highlighting the improvement in model accuracy and system response for wave compensation.
In spite of these developed control strategies, many low-cost Stewart platforms employ only kinematic controllers without taking into consideration the system dynamics which are crucial when handling changes in model parameters such as component masses and inertia.
Furthermore, validation of these methods is often restricted to simulation studies, while the limited experimental results that do exist are typically obtained using commercial platforms.

In this paper, we present the design and hardware/software realization of a low-cost Stewart platform prototype that combines off-the-shelf components with custom 3D-printed parts, achieving sufficient real-time performance of the platform for research and educational purposes. 
The prototype developed in this research includes a mechanical system model to characterize the kinematic and dynamic relationships between actuator displacements and platform motion, as well as a software system that integrates data acquisition, dynamic modeling, and real-time control elements of the platform. 
To evaluate the functionality of the prototype, a real-time controller is implemented that integrates feedback linearization with an LQR scheme based on feedback information obtained from an EKF-based state estimator to achieve robust trajectory tracking capability under noisy sensor measurement. 
In contrast with previous studies that focus mainly on theoretical modeling or simulation of isolated control implementations, this work provides a unified hardware–software framework that enables experimental validation of control strategies and state estimation algorithm on an affordable platform.  
The main contributions of this paper are summarized below.
\begin{enumerate}
    \item Develop a low-cost Stewart prototype that integrates off-the-shelf components with custom-designed parts.  
    \item Proposes an end-to-end design framework of the Stewart platform, developed entirely from scratch, covering mechanical, electronic, and software aspects of the platform.  
    \item Demonstrate experimental validation through the implementation of baseline control methods, demonstrating the functionality and applicability of the proposed platform.  
\end{enumerate}

The rest of the paper is organized as follows.
Section~\ref{sec:prototype} describes the developed Stewart platform prototype.
Section~\ref{section2} derives the kinematic and dynamic models of the platform, and states the objective of the control system design.
Section~\ref{sec:est-control} proposes a control framework that consists of an EKF-based estimator and a feedback linearization scheme.
Section~\ref{sec:sim-exp} presents detailed simulation and experimental results of the closed-loop system when subjected to varying desired poses.
Section~\ref{sec:conclusion} concludes the paper with remarks and plans for future work.


\section{Prototype Development and Description}
\label{sec:prototype}

\begin{figure}[!b]
\centering
\subfloat[Schematic and parts]{%
  \includegraphics[width=0.35\linewidth]{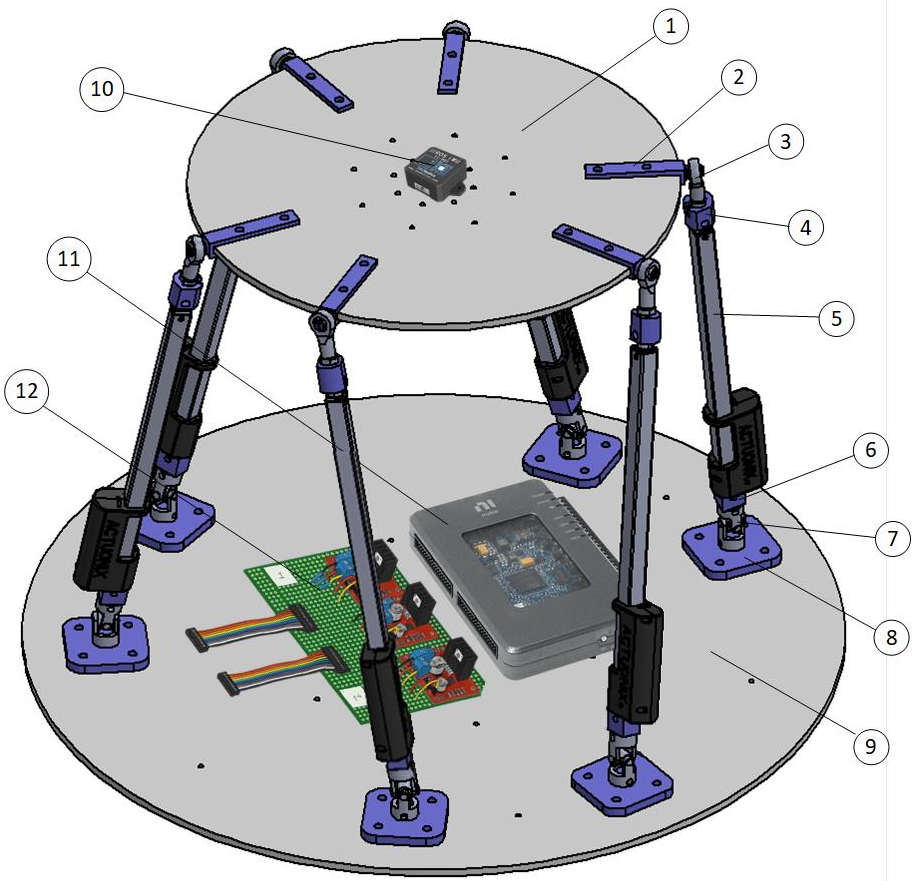}%
  \label{fig:stewart_platform}}
  \hspace{2cm}
\subfloat[Constructed prototype]{%
  \includegraphics[width=0.35\textwidth, height=5.2cm]{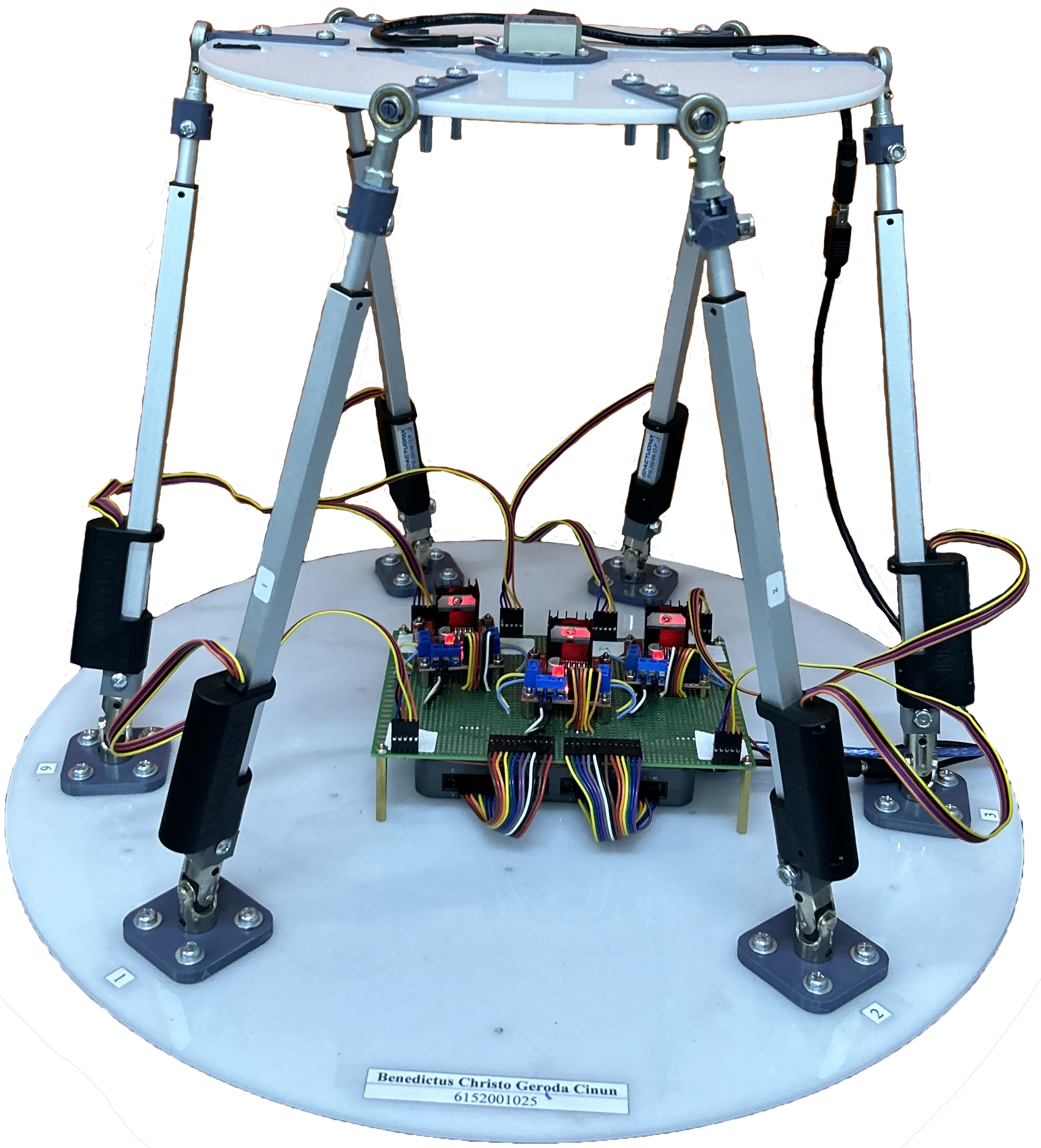}%
  \label{fig:real_SP1}
}%
\vspace{-3pt}
\caption{The constructed prototype of the Stewart platform.}
\end{figure}

The Stewart platform prototype discussed in this paper was developed using a combination of off-the-shelf and custom-made components to balance cost, functionality, and flexibility.
As illustrated in the Bill of Materials (BoM) in Table~\ref{tab:spending} of Appendix A, the total cost of our developed prototype is USD \$1,890.85, which is approximately 4.7 times cheaper than one of the cheapest commercially available Stewart platform product made by Acrome robotics.
Fig.~\ref{fig:stewart_platform} illustrates the design schematic, while Fig.~\ref{fig:real_SP1} shows the constructed prototype of the Stewart platform.
The circular shape platform \textcircled{\scriptsize 1} is made from 5 mm acrylic sheet, laser-cut with high precision, and serves as the end-effector with six degrees of freedom.
It is connected via a custom-made 3D-printed platform socket \textcircled{\scriptsize 2} to a POS 5 ASB bearing \textcircled{\scriptsize 3}, which functions as a passive spherical joint.
On the other hand, this bearing is also linked to the upper actuator adapter \textcircled{\scriptsize 4}, which is attached to the rod of the linear actuator \textcircled{\scriptsize 5}, acting as the primary driving mechanism of the system. 
The platform uses Actuonix P16-P linear actuators with a stroke of 200 mm, and each actuator is equipped with a built-in potentiometer that provides direct measurement of leg length.
Before its operation, each actuator undergoes a detailed manual calibration to ensure an accurate and almost identical operational characteristic with other actuators.
The lower part of the linear actuator is secured by an adapter \textcircled{\scriptsize 6}, which connects to a 10 mm universal joint \textcircled{\scriptsize 7}, further inserted into the base socket \textcircled{\scriptsize 8} mounted firmly on the circular shape base \textcircled{\scriptsize 9}, forming the fixed structural foundation.
\begin{figure}[!b]
\centering
\subfloat[Electrical schematic of PCB]{%
  \includegraphics[width=0.35\textwidth]{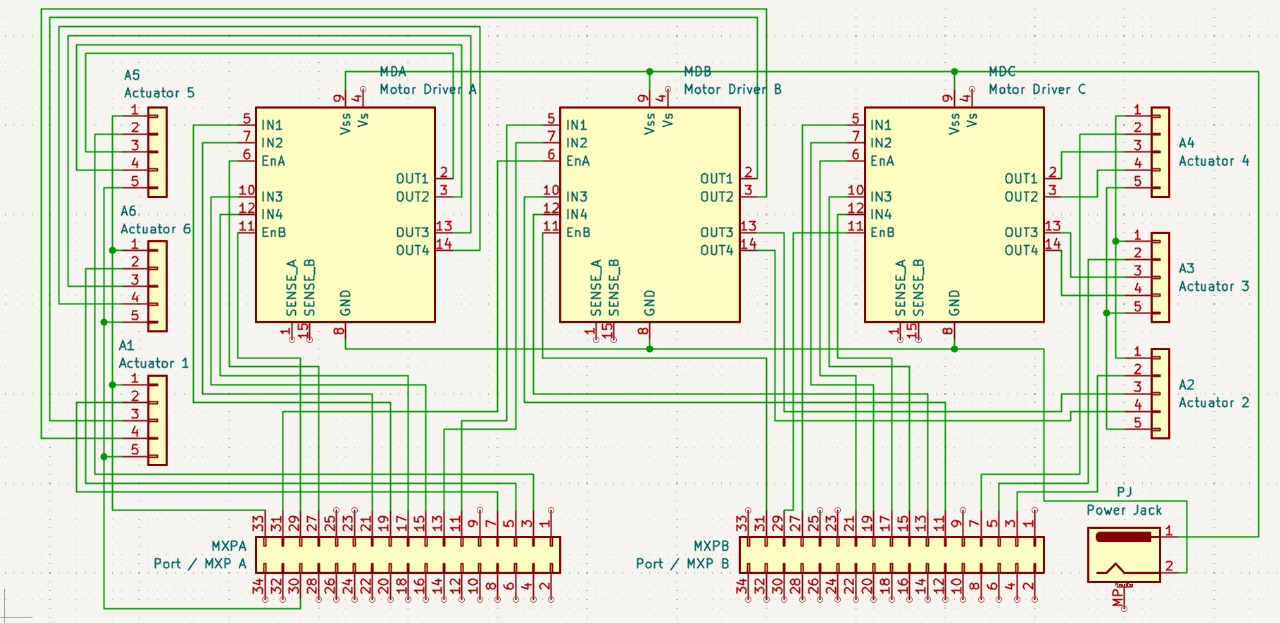}%
  \label{fig:pcb_schem}}
  \hspace{1cm}
\subfloat[Top-layer layout (front view) of the PCB]{%
  \includegraphics[width=0.35\textwidth, height = 2.7cm]{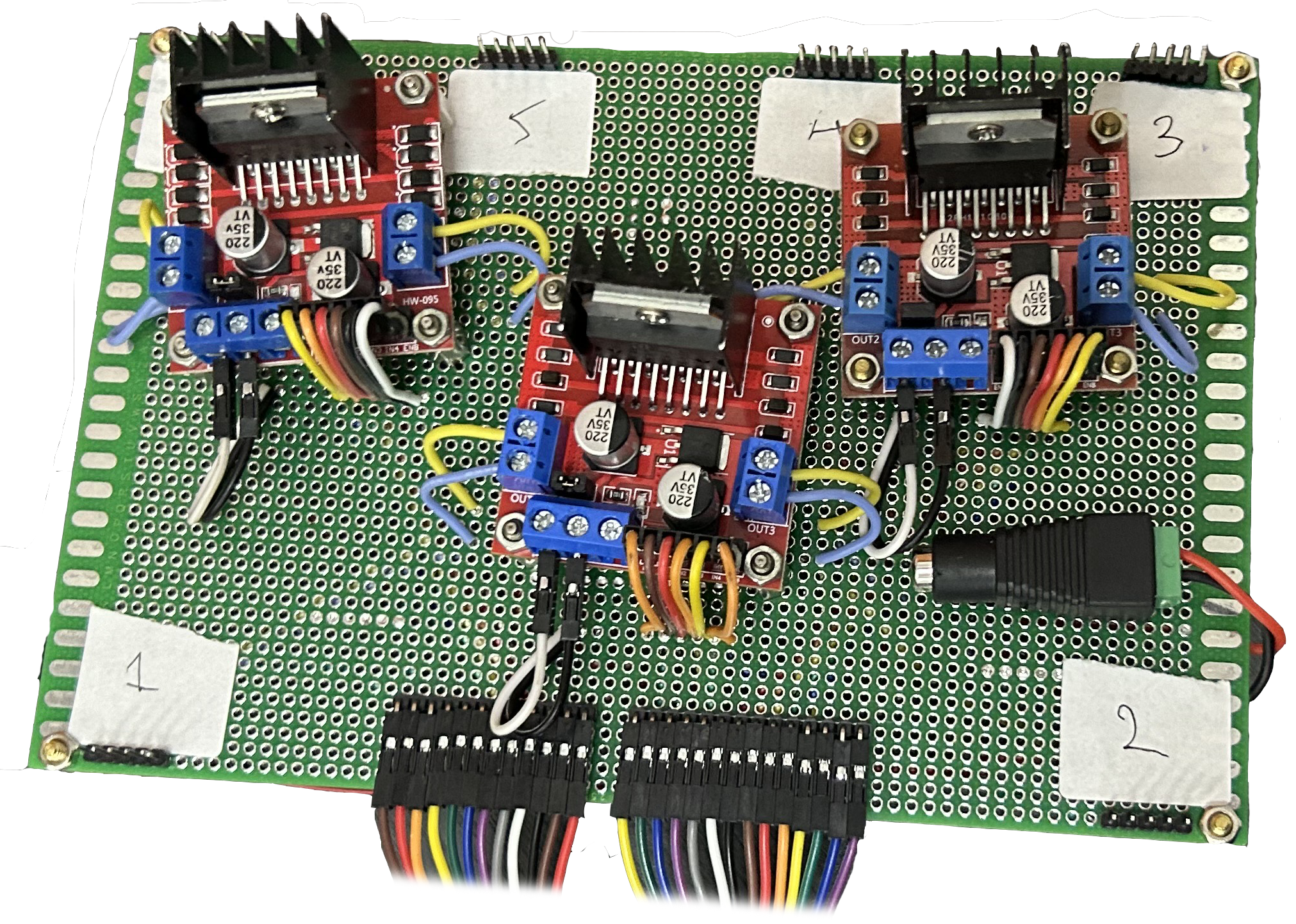}%
  \label{fig:pcb_real}
}%
\caption{Custom-designed PCB that interfaces the myRIO-1900 with motor driver and actuators.}
\end{figure}
In the center of the platform, an ROS IMU sensor \textcircled{\scriptsize 10}  is installed to measure the orientation and angular velocity of the platform in real time. 
The myRIO-1900 board \textcircled{\scriptsize 11} is used as the data acquisition module to process sensor data and execute control algorithms in coordination with a Personal Computer (PC).
This embedded device is chosen due to its versatility and connectivity with the LabVIEW software, which are suitable to be used for educational and research purposes.
The specifications of the myRIO-1900 board are summarized in Table \ref{tab:myrio-specs} in Appendix A \cite{ni_myrio1900_manual}.
The electronic system is supported by a custom-made Printed Circuit Board (PCB) \textcircled{\scriptsize 12}, which integrates motor drivers (L298N), power supply, and communication interface between myRIO and the six linear actuators.
The electrical schematic of the PCB is shown in Fig.~\ref{fig:pcb_schem} with its top layer layout as shown in Fig.~\ref{fig:pcb_real}.

\begin{figure}[!t]
    \centering
    \includegraphics[width=.65\linewidth]{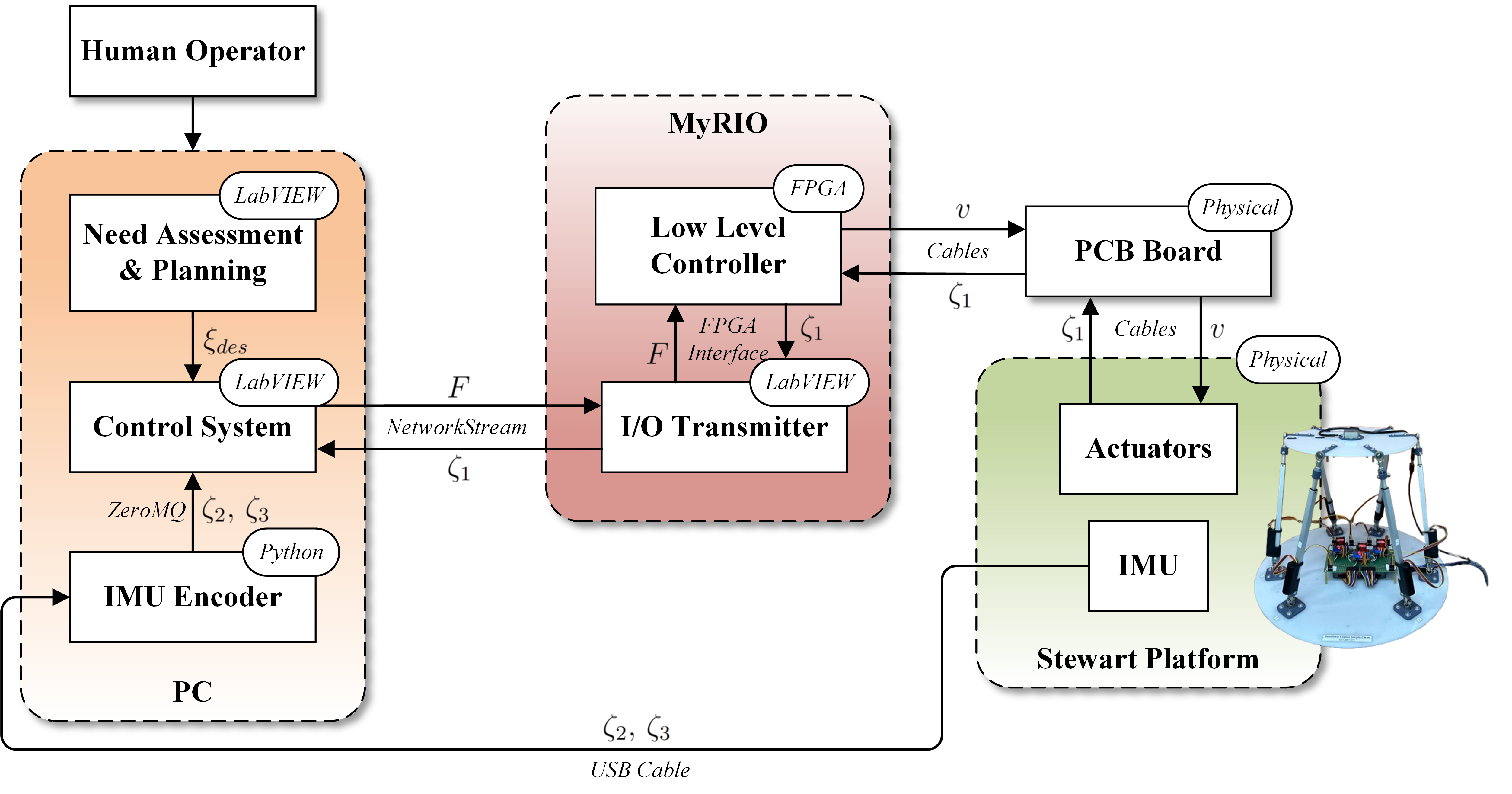}
    \caption{System configuration of the Stewart platform.}
    \label{fig:architecture}
\end{figure}

The communication architecture of the Stewart platform as illustrated in Fig.~\ref{fig:architecture} shows how relevant signals flow between sensors, actuators, and the main control system.
The myRIO-1900 executes the program that is developed in the LabVIEW software.
The processor handles communication with the PC and executes real-time control tasks, while the Field-Programmable Gate Arrays (FPGA) manages high-speed I/O with sensors and actuators.
The PC is responsible for high-level control, including trajectory planning and feedback processing, and has a Human-Machine Interface (HMI) that allows the operator to configure commands and monitor system states.
The PC also includes an IMU decoder to process and interpret the orientation and motion sensor data.

\begin{figure}[!t]
    \centering
    \includegraphics[width=0.3\linewidth]{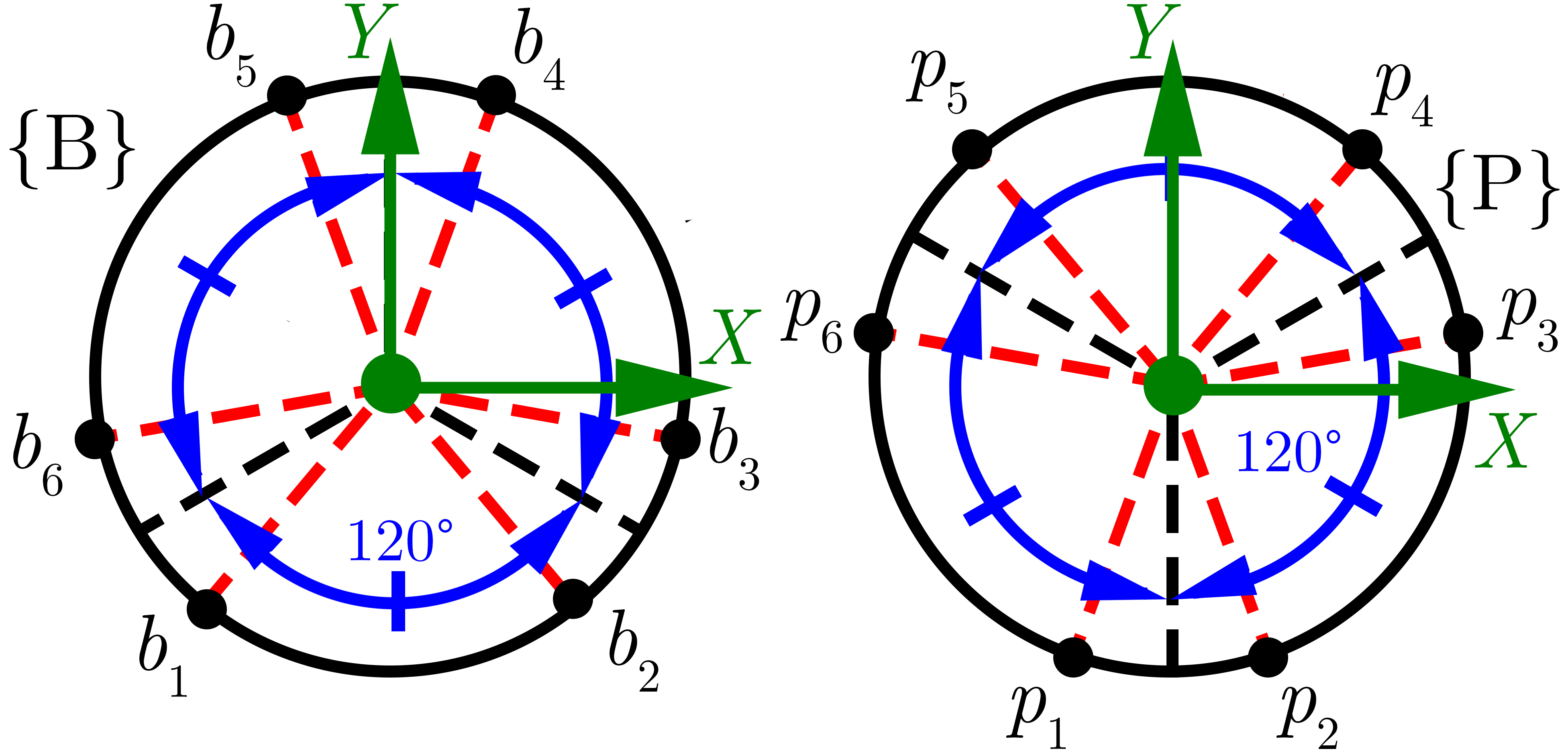}
    \caption{Joint configurations of the platform: joint $b_i$ positions on \{B\} (left) and joints $p_i$ position on \{P\} (right).}
    \label{fig:joint_pos}
\end{figure}

The prototype consists of a moving platform with radius $r_p = 16$ cm and a base with radius $r_b = 20$ cm. 
The joints $b_i$ and $p_i$ are defined in these elements and are symmetrically arranged in pairs, with the joints $b_i$ defined in the base frame $\{B\}$ and the joints $p_i$ in the platform frame $\{P\}$. 
As illustrated in Fig.~\ref{fig:joint_pos}, each of the three pairs of jonts is centered at 0$^0, 120^0$, and $240^0$ angles on $\{B\}$, and at $60^0, 180^0$, and $300^0$ angles on $\{P\}$, with a $\pm 20^0$ offset within each pair.
The circular acrylic platform has an effective radius of 16 cm (including joint connectors), a total mass of $m_p = 0.528$ kg, and a moment of inertia $\mathrm{I_p}$ as in \eqref{eqn:Ip_It_Ib}.
Each leg of the platform comprises a linear actuator with a bottom (static) part of mass $m_b = 0.1187$ kg and a top (moving) part of mass $m_t = 0.027$ kg. 
The center of mass of each part is assumed to be at its midpoint, giving distances of $l_b = 0.13861$ m and $l_t = 0.1$ m from the respective joints.
The corresponding moments of inertia $\mathrm{I_t}$ and $\mathrm{I_b}$ for the top and bottom parts, respectively, are specified as in \eqref{eqn:Ip_It_Ib}.
\begin{equation}
        \mathrm{I_p} = \noindent\begin{bmatrix}
        0.03 & 0.01 & 0.01 \\
        0.01 & 0.03 & 0.01 \\
        0.01 & 0.01 & 0.02
        \end{bmatrix}, \qquad
        \mathrm{I_t} = \noindent\begin{bmatrix}
\frac{1}{3} m_t l_t^2 & 0 & 0 \\
0 & \frac{1}{3} m_t l_t^2 & 0 \\
0 & 0 & 0
\end{bmatrix}, \qquad
\mathrm{I_b} = \noindent\begin{bmatrix}
\frac{1}{3} m_b l_b^2 & 0 & 0 \\
0 & \frac{1}{3} m_b l_b^2 & 0 \\
0 & 0 & 0
\end{bmatrix}.
        \label{eqn:Ip_It_Ib}
\end{equation}
%
%
%
%
\section{System Modeling and Problem Formulation}
\label{section2}
\begin{figure}[!t]
\centering
\subfloat[Kinematics diagram]{%
  \includegraphics[width=0.35\textwidth]{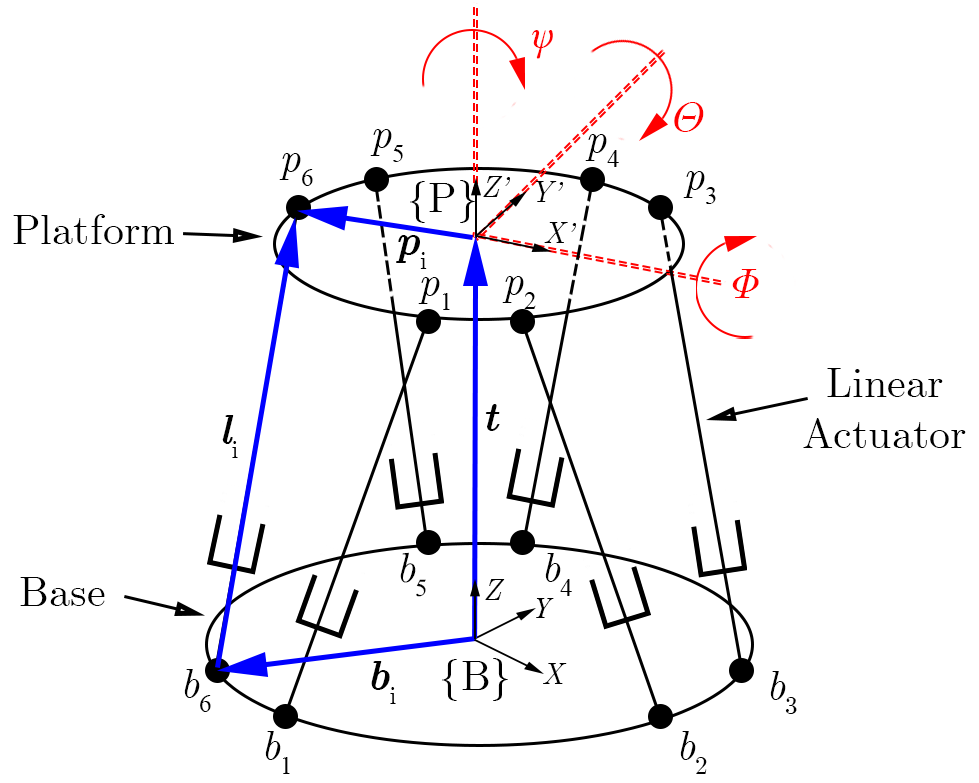}%
  \label{fig:schem_kin}}
  \hspace{2cm}
\subfloat[FBD of linear actuator]{%
  \includegraphics[width=0.25\linewidth]{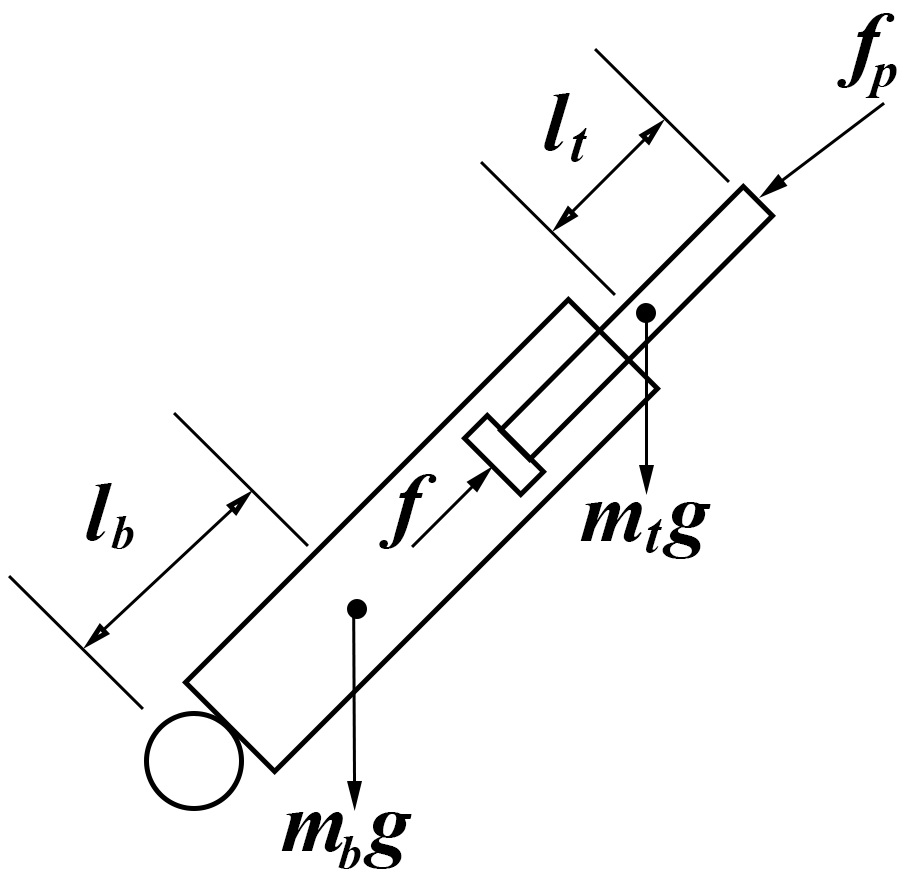}
  \label{fig:linact}
}%
\caption{Platform schematics used for kinematics and dynamics modeling.}
\end{figure}

Fig.~\ref{fig:schem_kin} shows the kinematic structure of the Stewart platform with three main structural elements: a fixed \textit{base} with an attached base frame \{B\}, a moving \textit{platform} with a platform frame \{P\}, and \textit{legs} formed by six linear \textit{actuators} that connect the base to the platform.
The free body diagram (FBD) of the linear actuator used for dynamics modeling is shown in Fig.~\ref{fig:linact}.
The actuator attachment points on the base are denoted as $b_i$ and those on the platform as $p_i$, where $i=1,\dots,6$.
These points serve as passive joints that mechanically connect the legs to the platform and the base, allowing relative motions to occur.
The joints in $b_i$ and $p_i$ are universal and spherical joints, respectively.

\subsection{Kinematic Model}

The structure of the Stewart platform enables six DoF motions of the platform with respect to (w.r.t.) the base, consisting of translational movements along the $X$, $Y$ and $Z$ axes, as well as rotational movements about these same axes.
The translation of $\{P\}$ w.r.t. $\{B\}$ is described by $t = [ X', Y', Z' ]^\intercal $, while the orientation is represented by the Euler angle vector $r = [\phi,\theta,\psi]^\intercal $.
Here, $\phi$ denotes roll (rotation about the $X$-axis), $\theta$ denotes pitch (rotation about the $Y$-axis), and $\psi$ denotes yaw (rotation about the $Z$-axis).
The vector of generalized pose of the Stewart platform is defined as
$q=[t,r]^\intercal =[ X', Y', Z', \phi,\theta,\psi]^\intercal  \in \mathbb{R}^{6}$.
The orientation of the platform is further described by the rotation matrix ${R} \in \mathrm{SO}(3)$, which maps the vectors from \{P\} to \{B\}.
The matrix ${R}$ is computed from the Euler angles $r$ using the ZYX (yaw–pitch–roll) convention ~\cite{niku2020introduction}.
The generalized velocity of the platform is defined as $\dot{q}=[\dot{t},\dot{r}]^\intercal $, where $\dot{t}=[X', Y', Z']^\intercal $ is the linear velocity of the platform, and $\dot{r}=[\dot{\phi},\dot{\theta},\dot{\psi}]^\intercal $ is the time derivatives of the Euler angles~\cite{csumnu2017simulation, yadavari2024addressing}.
The angular velocity expressed w.r.t. \{P\} is denoted by $\omega_p=[\omega_{px},\omega_{py},\omega_{pz}]^\intercal $ and is defined as follows.
\begin{align}\label{eq:omega_p}
{\omega}_p=\noindent\begin{bmatrix}
1 & 0 & -s_\theta \\
0 & c_\phi & c_\theta \, s_\phi \\
0 & -s_\phi & c_\theta \, c_\phi
\end{bmatrix}
\begin{bmatrix}
\dot{\phi} \\
\dot{\theta} \\
\dot{\psi}
\end{bmatrix}.
\end{align}
where $s_\theta := \sin(\theta)$ and $c_\phi := \cos(\phi)$.
The angular velocity in the frame $B$ can be rewritten as: $\tilde{\omega}={R}\tilde{\omega}_p{R}^\intercal,$
where $\tilde{\omega}$ and $\tilde{\omega}_p$ are the skew-symmetric matrices that correspond to the angular velocity vectors $\omega$ and ${\omega_p}$, respectively.

As shown in Fig.~\ref{fig:schem_kin}, ${b_i}$ denotes the position vector from the origin of $\{B\}$ to the base attachment point of the $i$-th leg, while ${p_i}$ denotes the corresponding vector from the origin of $\{P\}$ to the platform attachment point.
For a given platform pose $q$, the inverse kinematics equation which defines the necessary length that each leg must achieve to realize a given platform pose can be formulated and solved.
This involves using the known platform pose and the geometric configuration of the mechanism to determine the required leg lengths.
Specifically, the position and orientation of the platform, along with the fixed locations of the joints, are used to calculate the vector form of each leg.
Thus, the vector $l_i$ of the $i$-th leg with length $s_i=\|l_i\|$ can be expressed w.r.t. \{B\} as ~\cite{guo2006dynamic, liang2022kinematics}
\begin{equation}\label{eq:leglength}
    l_i = t + {R} \, {p}_i - {b_i},
\end{equation}

\subsection{Dynamics Model}

The dynamic model of the Stewart platform considers both the moving platform and the connecting legs.
The dynamics of the legs are formulated using Euler–Lagrange method, while those of the platform are modeled using Newton-Euler formulation~\cite{guo2006dynamic,bi_inverse_2014}.
The model is expressed in task space, allowing a direct relationship between the actuator forces and platform motion in terms of position and orientation.
This dynamic model is essential to accurately capture the coupled motion of the system and serves as the foundation for implementing model-based control strategies.


\subsubsection{Legs Dynamics Analysis}
In the prototype, each leg is realized using a linear actuator that functions as a prismatic joint, see Fig.~\ref{fig:linact}.
Each actuator is composed of two primary rigid components: a fixed section, referred to as the bottom part that is mounted to the base, and a movable section, referred to as the top part that is connected to the moving platform.
The bottom and top parts have masses of $m_b$ and $m_t$, respectively, and their centers of mass are located at distances $l_b$ and $l_t$ from the base and platform joints, respectively.
These components are subjected to external forces including gravitational force $g$, the actuator-generated force $f$, and the constraint force $f_p$ acting on the platform joint.

To determine the force $f_{p}$ on the $i$-th leg (defined as $f_{pi}$), Euler–Lagrange equation that takes into account the kinetic and potential energies of the actuator is used.
The resulting expression for the force $f_{pi}$ is defined as follows.
\begin{equation}
        f_{pi} = (\mathrm{M}_1 + \mathrm{M}_2)_i\noindent\begin{bmatrix}
            \mathrm{I}_{3\times 3} & {R}\tilde{p}_i^\intercal {R}^\intercal 
        \end{bmatrix}\ddot{q} + \mathrm{C_a}_i\noindent\begin{bmatrix}
            \mathrm{I}_{3\times 3} & {R}\tilde{p}_i^\intercal {R}^\intercal 
        \end{bmatrix}\dot{q} + (\mathrm{M}_1 + \mathrm{M}_2)_i\tilde{\omega}^2{R}p_i - (Q_f + Q_{m_tg} + Q_{m_bg})_i ,
    \label{eqn:fp_i}
\end{equation}
\noindent where $\tilde{{p}}_i$ is the skew-symmetric matrix of ${p}_i$, while the mass terms $\mathrm{M}_1$ and $\mathrm{M}_2$, the Coriolis term $\mathrm{C_a}$, and the generalized force components $Q_f$, $Q_{m_tg}$ and $Q_{m_bg}$ are given in the Appendix B.
The readers may also refer to~\cite{guo2006dynamic} for detailed derivation of \eqref{eqn:fp_i}.
In essence, \eqref{eqn:fp_i} relates the force applied by each actuator to the dynamic motion of the platform.




\subsubsection{Platform Dynamics Analysis}
The platform's dynamics is determined w.r.t. the control point of the platform.
This point does not always coincide with the origin of $\{P\}$, but can change depending on the position of the loads at its top. 
The vector of this control point w.r.t. $\{B\}$ can be written as follows.
\begin{align}
    q_c = t+{R}c_p,
\end{align}
\noindent where $c_p$ is the position vector of the control point w.r.t. \{P\}.
The acceleration of the control point $q_c$ satisfies \eqref{eq:qcddot}.
\noindent\begin{align}
    \ddot{q}_c = \noindent\begin{bmatrix}
        \mathrm{I} & {R}\tilde{c}_p^\intercal {R}^\intercal 
    \end{bmatrix}\ddot{q} + \tilde{\omega}^2{R}c_p.
    \label{eq:qcddot}
\end{align}
Assuming equilibrium forces are acting on the platform, Newton's law analysis on the platform gives the following.
\noindent
\noindent\begin{equation}
    m_p \ddot{q}_c = -\sum^{6}_{i=1}f_{pi} + m_pg,
    \label{eqn:platform_newton}
\end{equation}
\noindent where $m_p$ is the platform's mass.
By substituting \eqref{eq:qcddot} into \eqref{eqn:platform_newton}, the platform dynamics w.r.t. \{B\} satisfies
\noindent\begin{align}
    \noindent\begin{bmatrix}
        m_p \mathrm{I} & m_p{R}\tilde{c}_p^\intercal {R}^\intercal 
    \end{bmatrix} \ddot{q}  + m_p \tilde{\omega}^2{R}c_p= -\sum^{6}_{i=1}f_{pi} + m_pg.
    \label{eqn:platform_newton_base}
\end{align}

Combining the dynamic models of the legs and the platform, a compact expression of the dynamic model of the Stewart platform in the task space can be formulated as the following Equations of Motion (EoM)~\cite{guo2006dynamic, csumnu2017simulation}:
\noindent\begin{equation}
    \ddot{q} = \mathrm{M}^{-1}(q)\big(\mathrm{H}(q)\mathrm{F}-\mathrm{C}(q,\dot{q})\dot{q}-\mathrm{G}(q)\big),
    \label{eqn:stewart_dyn_equation}
\end{equation}
\noindent 
where $M(q) \in \mathbb{R}^{6 \times 6}$ is the inertia matrix, $\mathrm{C}(q, \dot{q}) \in \mathbb{R}^{6 \times 6}$ accounts for Coriolis and centrifugal effects, $\mathrm{G}(q) \in \mathbb{R}^{6}$ is the gravity vector, and $\mathrm{H}(q) \in \mathbb{R}^{6 \times 6}$ is the inverse transpose of the Jacobian matrix that maps the forces on the actuators to generalized forces in the task space.
Finally, $\mathrm{F} \in \mathbb{R}^{6}$ denotes the force inputs generated by the actuators.
Interested readers may refer to~\cite{guo2006dynamic} for a detailed derivation of the EoM in \eqref{eqn:stewart_dyn_equation} .

\subsection{Problem Formulation}
\label{sec:problem}
Given the EoM for the dynamics of the Stewart platform in \eqref{eqn:stewart_dyn_equation}, our objective is to construct a feedback controller that can move the platform from an initial pose to a desired final pose.
To this end, let $\xi = [q^\intercal ,\,\dot{q}^\intercal ]^\intercal $ be the vector of state variables of the platform.
One main challenge in designing a feedback controller for \eqref{eqn:stewart_dyn_equation} is that these state variables cannot be measured directly.
In this regard, an estimator should first be developed to provide estimates of the state variables based on measurement data obtained from available sensors. 
In this paper, the estimator is developed using the EKF algorithm, while the controller is developed using the combination of the feedback linearization scheme and the LQR control method.
Section \ref{sec:est-control} details these estimation and controller design.


\section{Estimation and Control Systems Design}
\label{sec:est-control}
This section presents the design of a controller for the developed Stewart platform using the integration of a feedback linearization scheme and the LQR method.
Since the platform state variables cannot be measured directly by the available sensors, an EKF algorithm is designed to calculate the estimate of these state variables and then used in the proposed feedback control strategy.
The integration of the feedback controller and the EKF-based estimation scheme helps to achieve more accurate motion execution with low computational demand.
This makes the framework suitable for real-time deployment on the Stewart platform prototype.

\subsection{Feedback Linearization Control Scheme}
\label{section3.1}
To achieve a precise motion of the Stewart platform, a practical control strategy based on the feedback linearization scheme is implemented.
This scheme essentially transforms the dynamics into a linear system model that is suitable for the linear control design framework~\cite{isidori1995nonlinear}.
For the EoM of the Stewart platform in \eqref{eqn:stewart_dyn_equation}, a feedback linearizing force control input $\mathrm{F}$ of the form \eqref{eq:force} is designed to cancel the system nonlinearities:
\noindent\begin{equation}\label{eq:force}
    \mathrm{F} := \mathrm{H}^{-1}(q)\left(\mathrm{M}(q){{u}} + \mathrm{C}(q,{\dot{q}}){\dot{q}} + \mathrm{G}(q)\right),
\end{equation}
where ${u}$ is a virtual control input that must be designed for the resulting linear system. 
Substituting \eqref{eq:force} into \eqref{eqn:stewart_dyn_equation} results in a double integrator model $\ddot{q}={u}$ which can be written as the following linear time-invariant (LTI) system model.
\noindent\begin{align}
    \noindent\begin{bmatrix}
        \dot{q} \\
        \ddot{q}
    \end{bmatrix} &= 
    \noindent\begin{bmatrix}
        {0}_{3\times 3} & \mathrm{I}_{3\times 3} \\
        {0}_{3\times 3} & {0}_{3\times 3}
    \end{bmatrix}
    \noindent\begin{bmatrix}
        q \\
        \dot{q}
    \end{bmatrix} + 
    \noindent\begin{bmatrix}
        {0}_{3\times 3}  \\
        \mathrm{I}_{3\times 3} 
    \end{bmatrix}
    {u}.
    \label{eq:state_space}
\end{align}
Define $\xi=[q,\dot{q}]^\intercal $, then \eqref{eq:state_space} can be rewritten as the following compact LTI system model.
\noindent\begin{equation}
    \dot{{\xi}} = {A}{\xi} + {B}{u}.
    \label{eq:ss-lti}
\end{equation}

The LTI system model \eqref{eq:ss-lti} allows for the design of a linear feedback control law of the form
\noindent\begin{equation}
    {u} = -{K} \{{\xi}-{{\xi}}_{des}\},
    \label{eq:lqrgain}
\end{equation}
where ${\xi}_{des}$ is the desired state variable and $K$ is the feedback control gain. 
In this paper, the feedback control $u$ is designed using LQR scheme to ensure the minimization of a quadratic performance index $J(\xi,u)$ as follows~\cite{aastrom2021feedback}.
\noindent\begin{equation}
   u:=\arg\min J(\xi,u), \;\text{ where }\; J = \int_{0}^{t} ({\xi}^\intercal {N}{\xi} + {u}^\intercal {O}{u})dt,
    \label{eqn:perf_index}
\end{equation}
where $N\in \mathbb{R}^{12 \times 12}$ and ${O} \in \mathbb{R}^{6 \times 6}$ are the state and control input weighting matrices, respectively, which directly influence the system's behavior: ${N}$ penalizes deviations from the desired state to achieve accurate tracking, while ${O}$ regulates the control effort to ensure smooth operation by preventing excessive actuator efforts.
The optimal feedback gain ${K}$ in \eqref{eq:lqrgain} is calculated as $K = O^{-1}{B}^\intercal {D}$, where $D$ is a positive semidefinite matrix which satisfies the continuous-time algebraic Riccati equation of the form:
${A}^\intercal {D} + {D}{A} - {D}{B}{R}^{-1}{B}^\intercal {D} + N = {0}$.

\subsection{EKF-Based State Estimation Scheme}
\label{section3.2}
The implementation of control methods in Section \ref{section3.1} requires feedback information about the state variables from sensors.
However, the Stewart platform prototype developed in this research utilizes three types of measurement that do not directly measure the state variables. 
The first measurement is obtained from a built-in encoder embedded in each linear actuator which measures the length of each leg on the platform.
The encoder measurement provides indirect information on the pose of the platform, which can be calculated using the inverse kinematics relationship in \eqref{eq:inversekine}. 
The second measurement is given by an inertial measurement unit (IMU) mounted on the moving platform to provide real-time rotational data which capture the orientation state $r$. 
The last measurement is also obtained from the IMU and provides $\omega_p$ in \eqref{eq:omega_p}.
In this regard, the following function $\gamma$ is used as the output model.
\noindent\begin{align} \label{eq:measurement}
    \gamma &= [s_1,s_2,s_3,s_4,s_5,s_6,\phi,\theta,\psi,\omega_{px},\omega_{py,}\omega_{pz}]^\intercal +v,
\end{align}
where $v\sim\mathcal{N}(0, V)\in \mathbb{R}^{12}$ is a Gaussian noise vector with mean zero and covariance matrix of \(V \in \mathbb{R}^{12\times12}\).

Using \eqref{eq:measurement}, an EKF algorithm is developed to estimate the state variables $\hat{\xi}$ that will used for control algorithm implementation in Section \ref{section3.1}.
The EKF is developed using the discrete-time version of the system model in \eqref{eq:state_space} and the output measurement in \eqref{eq:measurement}.
Using the forward Euler method for a sampling period of \(\Delta t\), the discretization of \eqref{eq:state_space} at each discrete time instant \(k \in \mathbb{N}\) of the following form is considered.
\noindent\begin{equation}
    \noindent\begin{aligned}\label{eq:discrete_state_space}
        \noindent\begin{bmatrix}
        {q_{k{\mid} k{-}1}} \\
        \dot{q}_{k{\mid} k{-}1}
    \end{bmatrix} &= 
    \noindent\begin{bmatrix}
        \mathrm{I}_{3\times 3} & \Delta t\,\mathrm{I}_{3\times 3} \\
        {0}_{3\times 3} & \mathrm{I}_{3\times 3}
    \end{bmatrix}
    \noindent\begin{bmatrix}
        q_{k{-}1{\mid} k{-}1} \\
        \dot{q}_{k{-}1{\mid} k{-}1}
    \end{bmatrix} + 
    \noindent\begin{bmatrix}
        {0}_{3\times 3}  \\
        \Delta t\,\mathrm{I}_{3\times 3} 
    \end{bmatrix}
    {u}_{k{-}1} +w_k,\\
    \hat{{\xi}}_{k{\mid} k{-}1} &= {A}_d\,\hat{{\xi}}_{k{-}1{\mid} k{-}1} + {B}_d\,{u}_{k{-}1}+w_k,
    \end{aligned}
\end{equation}
where $w_k \in \mathbb{R}^{12} \sim \mathcal{N}(0, W)$ is an asumed Gaussian process noise with mean zero and covariance matrix of \(W \in \mathbb{R}^{12\times12}\).
The prediction model in \eqref{eq:discrete_state_space} essentially models the prior state estimate \( \hat{\xi}_{k|k{-}1} \) as the function of the previous state estimate \( \hat{\xi}_{k{-}1|k{-}1} \) and the control input \( u_{k{-}1} \). 
The implementation of the EKF also uses the Jacobian matrix \(\mathrm{\Gamma}_k\) of the measurement function \eqref{eq:measurement}, 
evaluated in the forecast state estimate \(\hat{\xi}_{k{\mid} k{-}1}\) as follows
\noindent\begin{align}
    \mathrm{\Gamma}_k = \left.\dfrac{\partial {\gamma}}{\partial {\xi}}\right|_{\xi=\hat{\xi}_{k{\mid} k{-}1}}
    \label{eq:measurement_jacobian_matrix}.
\end{align}

\begin{algorithm}[!b]
\caption{Pseudo code of the EKF.}
\noindent\begin{algorithmic}[1]
\Procedure{Initialization}{}
    \State At $k=0$, initialize $\hat{{\xi}}_{0{\mid} 0}$, ${P}_{0{\mid} 0}$
    \State \textbf{return} $\hat{{\xi}}_{0{\mid} 0},\; {P}_{0{\mid} 0}$
\EndProcedure
At each $k$, repeat the following procedures:
\Procedure{Prediction}{}
    \State Compute forecast PDF data ${\xi}_{k{\mid} k{-}1}$ and ${P}_{k{\mid} k{-}1}$ using \eqref{eq:forecast}
    \State \textbf{return} $\hat{{\xi}}_{k{\mid} k{-}1},\; {P}_{k{\mid} k{-}1}$
\EndProcedure
\Procedure{Measurement-update}{}
    \State Compute Jacobian matrix $\mathrm{\Gamma}_k$ in \eqref{eq:measurement_jacobian_matrix}, and construct the innovation vector $\tilde{{z}}_{k} = {z}_{k} - \left.{\gamma}\right|_{\xi=\hat{{\xi}}_{k{\mid} k{-}1}}$
    \State Compute posterior PDF data ${\xi}_{k{\mid} k}$ and ${P}_{k{\mid} k}$ using \eqref{eq:posterior_PDF}
    \State \textbf{return} $\hat{{\xi}}_{k{\mid} k},\; {P}_{k{\mid} k}$
\EndProcedure

\end{algorithmic}
\label{alg:basic_ekf}
\end{algorithm}

Algorithm \ref{alg:basic_ekf} summarizes the EKF that is used to estimate the Probability Density Function (PDF) of the state estimate $\hat{{\xi}}$ and the state covariance matrix ${P}\in \mathbb{R}^{12\times 12}$.
The EKF method consists of two stages, namely \textit{Prediction} and \textit{Measurement-Update}.
In the \textit{Prediction} stage, the PDF is propagated forward from the prior PDF according to 
\noindent\begin{equation}
    \hat{{\xi}}_{k{\mid} k{-}1} = {A}_d\,\hat{{\xi}}_{k{-}1{\mid} k{-}1} + {B}_d\,{u}_{k{-}1},\qquad
    {P}_{k{\mid} k{-}1} = {A}_d \, {P}_{k{-}1{\mid} k{-}1} \, {A}_d^\intercal  + {V}.
    \label{eq:forecast}
\end{equation}
The \textit{Measurement-update} stage then updates the forecast PDF into a posterior PDF by incorporating the latest observation $z_k$ provided in the output vector.
The innovation vector $\tilde{{z}}_{k} = {z}_{k} - \left.{\gamma}\right|_{\xi=\hat{{\xi}}_{k{\mid} k{-}1}}$ is constructed and defined in this step as the difference between the output vector \({z}_k\) and the predicted output measurement evaluated w.r.t. $\hat{{\xi}}_{k {\mid} k{-}1}$.
Finally, the posterior state estimate \(\hat{{\xi}}_{k{\mid} k}\) and the posterior covariance \({P}_{k{\mid} k}\) can be calculated using \eqref{eq:posterior_PDF}.
\noindent\begin{equation}
    \hat{{\xi}}_{k{\mid} k} = \hat{{\xi}}_{k{\mid} k{-}1} + {P}_{k{\mid} k{-}1} \, \mathrm{\Gamma}_{k}{}^\intercal  \, \left( \mathrm{\Gamma}_{k} \, {P}_{k{\mid} k{-}1} \, \mathrm{\Gamma}_{k}{}^\intercal  + {W} \right)^{-1} \, \tilde{{z}}_{k},\qquad
    {P}_{k{\mid} k} = \left(\mathrm{I} - {K}_{k} \, \mathrm{\Gamma}_{k}\right) {P}_{k{\mid} k{-}1}.
    \label{eq:posterior_PDF}
\end{equation}
Algorithm \ref{alg:basic_ekf} begins with an \textit{Initialization} phase where the initial state \(\hat{{\xi}}_{0{\mid} 0}\) is estimated from encoder data using numerical forward kinematics~\cite{FKgithub}, and the initial covariance \({P}_{0{\mid} 0}\) is chosen to be large enough to reflect initial uncertainty.

\subsection{Overall Estimation \& Control Systems Architecture}
\noindent\begin{figure}[!t]
    \centering
    \includegraphics[width=.45\linewidth]{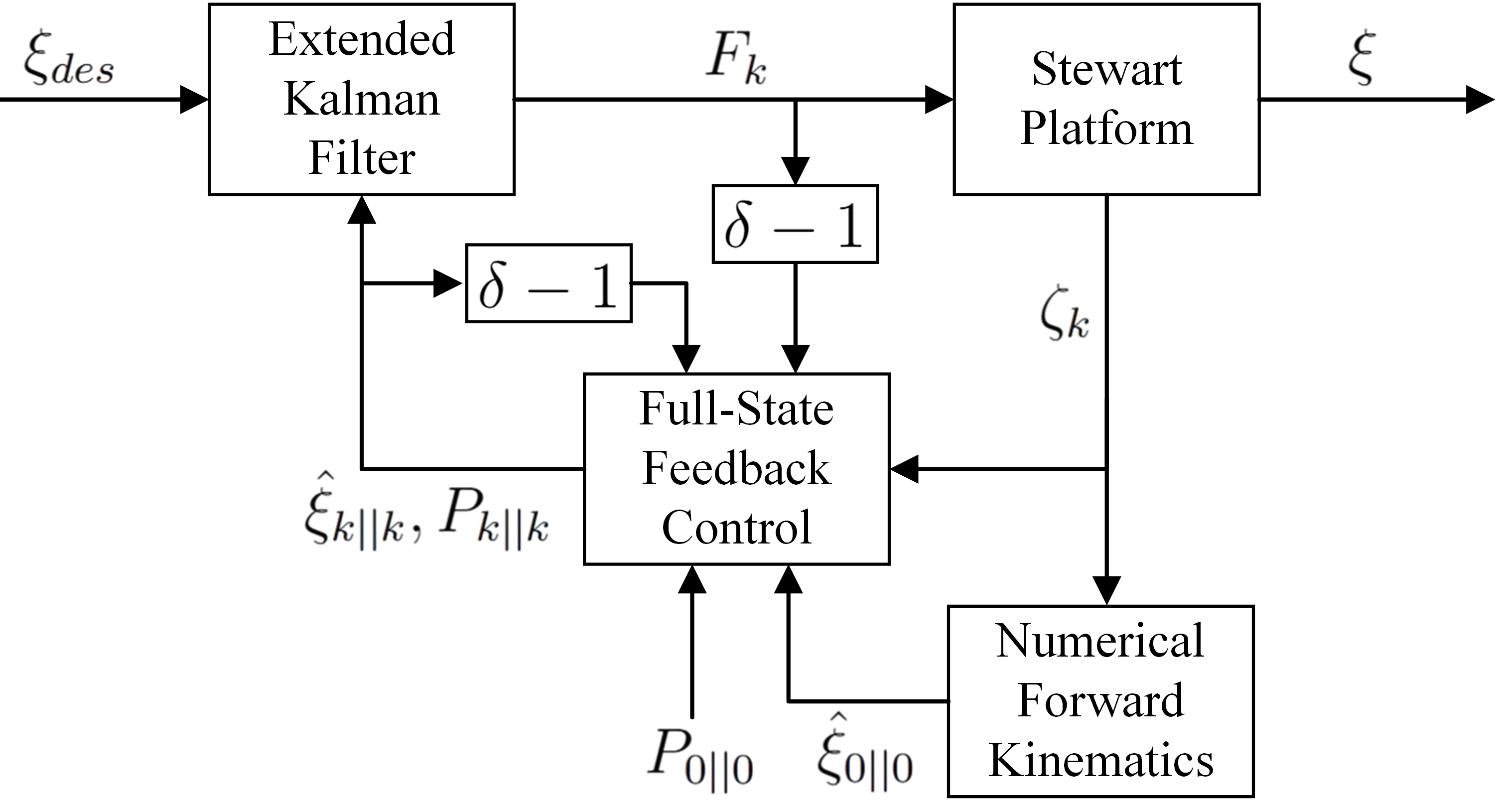}
    \caption{Block diagram of estimation and control.}
    \label{fig:overall_control_system}
\end{figure}
The overall architecture of the estimation and control systems for the Stewart platform is shown in Fig.~\ref{fig:overall_control_system}.
In this figure, the desired state variable ${\xi}_{des}$ serves as the reference input to the feedback control module, which in turn computes the desired forces $\mathrm{F}$ that drive the current vector of state variables of the Stewart platform to the desired one.
The estimate of the state variables used in the controller design is obtained using the EKF module based on measurement data ${z}_k$ obtained at the discrete time step $k$ from the leg potentiometers and the IMU sensor.
The EKF also incorporates the initial state estimates $\hat{{\xi}}_{0|0}$ and their covariance ${P}_{0|0}$ calculated using the combination of the numerical forward kinematics method and the measured encoder data~\cite{FKgithub}.
The posterior estimates of the state variables $\hat{{\xi}}_{k|k}$ and their covariance ${P}_{k|k}$ from the EKF are then fed back to the controller module.
The proposed estimation and feedback loop allow the controller to adjust its commands based on the real-time estimate of the state of the platform, which eventually enables robust and accurate control action of the platform motion.

%
%
%
%
\section{Simulation and Experimental Results}
\label{sec:sim-exp}
This section presents the simulation and experimental results of the implementation of the proposed feedback control scheme on the Stewart platform's model and prototype.
The initial condition of the platform is set to be:
$\xi_0 = [0, 0, 0.32, 0, 0, 0, 0, 0, 0, 0, 0, 0]^\intercal$, making the platform horizontally parallel and centered to the base.
The performance of the controller is tested under two desired motion scenarios of the platform, namely (a) \textit{step motion} and (b) \textit{sinusoidal motion}. 
The \textit{step motion} scenario was conducted for a duration of 60 seconds, while the \textit{sinusoidal motion} scenario lasted 20 seconds.
Both scenarios used the same sampling interval of 0.01 seconds.
The \textit{step motion} scenario aims to evaluate the platform's ability to track static desired states as defined in \eqref{eq:stepscenarios}.
\noindent\begin{equation}\label{eq:stepscenarios}
    q_{des} = [x_{des}, y_{des}, z_{des}, \phi_{des}, \theta_{des}, \psi_{des}]^\intercal, \qquad
    \dot{q}_{des} = [0, 0, 0, 0, 0, 0]^\intercal .
\end{equation}
where the other components of $q_{des}$ are defined over time $t$ as follows.
\noindent\begin{equation*}
\noindent\begin{aligned}
    x_{des} &=
\noindent\begin{cases}
    0.075 & \text{if } 10 \le t < 20 \\
    0 & \text{otherwise}
\end{cases},& 
y_{des} &=
\noindent\begin{cases}
    0.075 & \text{if } 20 \le t < 30 \\
    0 & \text{otherwise}
\end{cases},&
z_{des} &= 0.4,\\
\phi_{des} &=
\noindent\begin{cases}
    0.15 & \text{if } 30 \le t < 40 \\
    0 & \text{otherwise}
\end{cases},&
\theta_{des} &=
\noindent\begin{cases}
    0.15 & \text{if } 40 \le t < 50 \\
    0 & \text{otherwise}
\end{cases}
,&
\psi_{des} &=
\noindent\begin{cases}
    0.15 & \text{if } 50 \le t \le 60 \\
    0 & \text{otherwise}
\end{cases},
\end{aligned}
\end{equation*}
The sinusoidal motion scenario is designed to evaluate the platform's ability to track a dynamic trajectory in \eqref{eq:dancingscenarios}.
\begin{equation}
        q_{des} = [0,0,0.4,0.1\,s_t,0.1\,c_t,0]^\intercal,  \qquad
        \dot{q}_{des} = [0,0,0.4,0.1\,c_t,-0.1\,s_t,0]^\intercal  .
    \label{eq:dancingscenarios}
\end{equation}
A video demonstration of the presented results is available at~\cite{video-demo}.

\subsection{Simulation Results}
For the implementation of the EKF , the covariance matrices of the process and the measurement noises are set to
$ V = \texttt{diag}(1, 1, 1, 1, 1, 1, 5, 5, 5, 5, 5, 5)$ and $W = \texttt{diag}(10, 10, 10, 10, 10, 10, 1, 1, 1, 3, 3, 3)$, respectively. The initial value of the error covariance matrix is set to $P_0 = 10^{-2}\texttt{diag}(100, 100, 100, 100, 100, 100, 1, 1, 1, 1, 1, 1)$.
For the LQR controller, the state weighting matrix is chosen as
$N=\texttt{diag}(30, 30, 5, 30, 30, 200, 3, 3, 1, 3, 3, 20)$,
while the control input weighting matrix is set to $O = 10\mathrm{I}_6$.
These parameters were empirically tuned through several simulation trials to achieve desirable estimation and tracking performance.
For the estimation problem, the objective is to ensure the convergence of the state estimation error $e_l$ according to \eqref{eq:localization}.
    \noindent\begin{align}\label{eq:localization}
    \lim _{t \rightarrow \infty}e_l \leq r_{l}, \quad e_l=\left\|\hat{{\xi}}-{\xi}\right\|.
    \end{align}
For the control problem, the objective is to ensure the convergence of the tracking error $e_t$ according to \eqref{eq:tracking}.
    \noindent\begin{align}\label{eq:tracking}
    \lim _{t \rightarrow \infty} e_t\leq r_{t}, \quad e_t = \left\|{\xi}-{\xi}_{des}\right\|.
    \end{align}
In \eqref{eq:localization}-\eqref{eq:tracking}, $r_{l}$ and $r_{t}$ are some specified positive constants that were set in the simulations to be $r_t=0.02$ and $r_l=0.01$.

\begin{figure}[!t]
    \centering
        \centering
        \includegraphics[width=.9\linewidth]{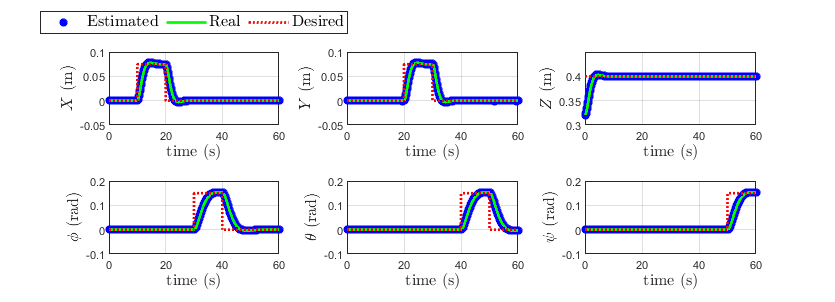}
        \vspace{-5pt}
        \caption{Simulated position variables of the \textit{step motion} scenario.}
       \label{fig:Sim_step_pos}
\end{figure}
\begin{figure}[!t]
        \vspace{-10pt}
    \centering
        \centering
        \includegraphics[width=.9\linewidth]{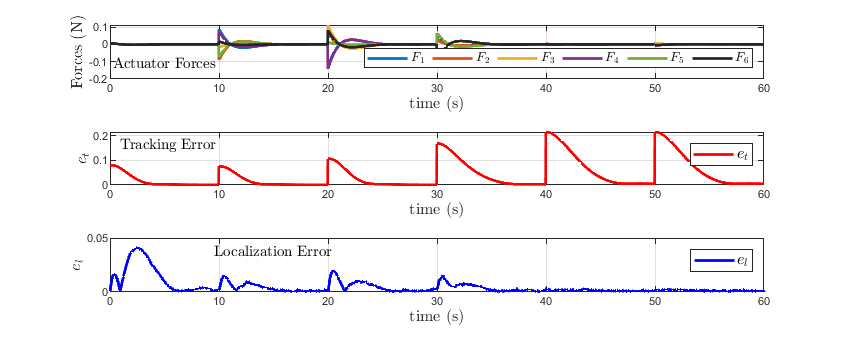}
        \vspace{-5pt}
        \caption{Simulated force control input of the \textit{step motion} scenario.}
       \label{fig:Sim_step_forces}
\end{figure}

\subsubsection{Simulation Result of \textit{Step Motion Scenario}}

The trajectories of the position of the platform for this scenario are shown in Fig.~\ref{fig:Sim_step_pos}, and the corresponding control input and estimation error are shown in Fig.~\ref{fig:Sim_step_forces}. 
In these figures, the actual states are shown by the green lines, whereas the estimated states are plotted in blue dots.
The close similarity between the actual and estimated states confirms the effectiveness of the proposed estimation scheme in capturing the system dynamics.
As shown in Fig.~\ref{fig:Sim_step_pos}, the calculated force control input drives the system to follow the desired pose.
Fig.~\ref{fig:Sim_step_forces} shows that the estimation error $e_l$ decreases as the state variables reach the desired values with a final value of approximately $0.0012$.
In addition, the tracking error $e_t$ also decreases and eventually converges to values within the acceptable  bound.
These results demonstrate that the proposed scheme can be applied to the static desired trajectory case.




\subsubsection{Simulation Result of \textit{Sinusoidal Motion} Scenario}

\begin{figure}[!t]
    \centering
        \centering
        \includegraphics[width=.9\linewidth]{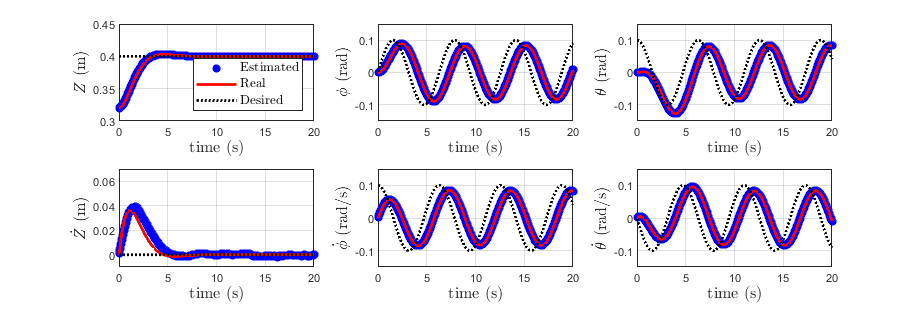}
        \caption{Simulated position variables of the \textit{sinusoidal motion} scenario.}
       \label{fig:State_sim_dancing}
       \vspace{-5pt}
\end{figure}

\begin{figure}[!t]
       \vspace{-10pt}
    \centering
        \centering
        \includegraphics[width=.9\linewidth]{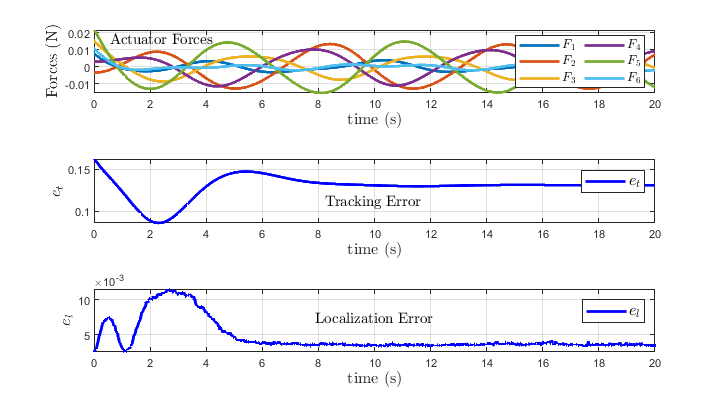}
       \vspace{-5pt}
        \caption{Simulated force control input of the \textit{sinusoidal motion} scenario.}
       \label{fig:Sim_dancing_forces}
\end{figure}

The state trajectories of the platform for this scenario are shown in Fig.~\ref{fig:State_sim_dancing}, while the corresponding control inputs and the estimation error are shown in Fig.~\ref{fig:Sim_dancing_forces}. 
The actual states variables are shown as the green lines, while the estimated values are plotted as a blue dot line.
The estimation scheme effectively captures the dynamics of the system, as evidenced by the close alignment between the estimated and actual states. 
Furthermore, the plots of $\psi$, $\theta$, and velocity in Fig.~\ref{fig:State_sim_dancing} demonstrate that the generated force control input successfully drives the platform to the desired pose.
In general, these results demonstrate the effectiveness of the proposed estimation and control method for a time-varying reference trajectory.
Compared to the \textit{step motion} scenario, the tracking error in the \textit{sinusoidal motion} exhibits smoother variations and eventually converges to approximately $0.13$.
Although this value exceeds the tolerance defined in \eqref{eq:tracking}, it is attributed to the continuously changing desired states and the accumulation of errors in all the states of the system.
Moreover, the estimation error also converges to approximately $0.0037$, which is well below the tolerance value of $r_l=0.01$.
From these results, it can be concluded that the platform can keep up with varying desired poses using the proposed estimation and control methods.



\subsection{Experimental Results}
\label{sec:experiment}

Due to the inherent challenges in directly measuring the true state variables in real-time, the experimental results are primarily compared against the outcomes of similarly tuned simulations.
This comparative analysis will allow for an assessment of the prototype's performance and the effectiveness of the proposed algorithms in a real-world scenarios, despite the absence of a perfectly known ground truth for the platform's state.
In particular, the main control objective in the experiment is to achieve the condition below: 
    \noindent\begin{align}\label{eq:ecs}
    \lim _{t \rightarrow \infty}e_{cs} \leq r_{\mathrm{cs}}, \quad e_{cs}=\left\|{\xi}_{des}-\hat{{\xi}}\right\|,
    \end{align}
where $r_{cs}$ is a specified positive constant.


\subsubsection{Experimental Result of \textit{Step Motion} Scenario}

\begin{figure}[!t]
    \centering
        \centering
        \includegraphics[width=.9\linewidth]{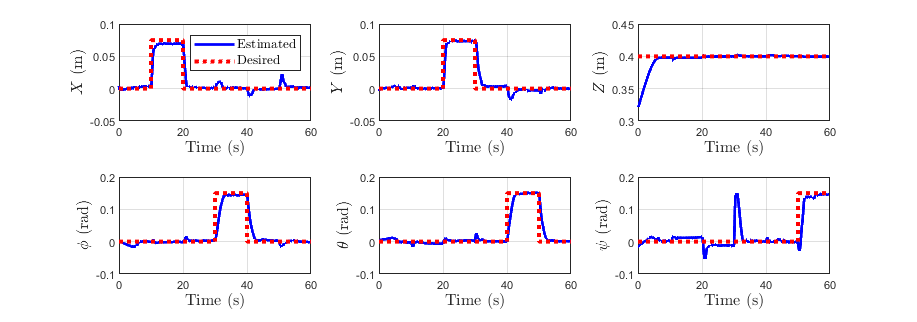}
        \vspace{-5pt}
        \caption{Experimental position variables of the \textit{step motion} scenario.}
       \label{fig:Exp_step_pos}
\end{figure}

\begin{figure}[!t]
\vspace{-10pt}
    \centering
        \centering
        \includegraphics[width=.9\linewidth]{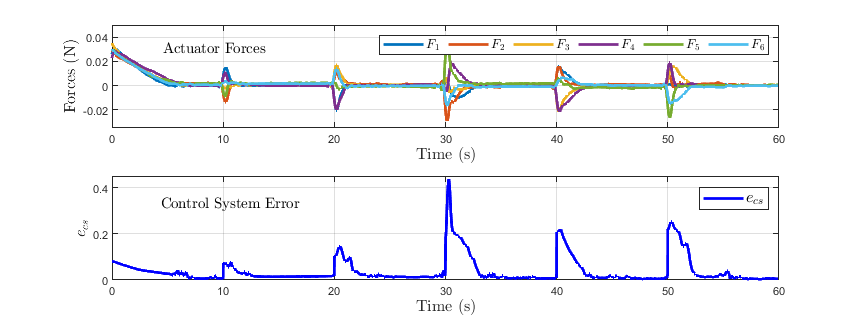}
        \vspace{-5pt}
        \caption{Experimental force control input of the \textit{step motion} scenario.}
       \label{fig:Exp_step_force}
\end{figure}

The position trajectories of the platform in the experiment are shown in Fig.~\ref{fig:Exp_step_pos}.
In both figures, the estimated states are shown as blue lines, while the desired trajectories are shown as purple dotted lines.
The generated force control inputs and the tracking error are shown in Fig.~\ref{fig:Exp_step_force}.
These figures show that the platform's position and orientation estimates closely follow the desired values, indicating the good performance of the proposed EKF-based state estimation scheme.
The results show that the platform can follow changes in the trajectory along the linear axes.
Although the $X$ and $Y$ axes exhibit noticeable fluctuations, the estimated states still capture the general shape and timing of the desired trajectories.
Although the $Z$ axis shows only minor fluctuations, those in the $X$ and $Y$ directions may appear more pronounced due to differences in the plot scale.
Small disturbances are also observed in the platform’s orientation.
These fluctuations are expected in experimental settings and do not significantly impact overall trajectory tracking performance.

The estimated velocity of the platform shows more noticeable fluctuations compared to the position estimates.
Although these fluctuations are relatively minor in the linear velocity components ($\dot{X}$, $\dot{Y}$, $\dot{Z}$), they become more pronounced in the angular velocity estimates ($\dot{\phi}$, $\dot{\theta}$, $\dot{\psi}$).
At certain time intervals, the velocity profiles exhibit overshoot, which is primarily due to rapid changes in the desired position.
These overshoots occur as the platform moves toward the new desired pose before coming to rest and stabilizing.
This behavior is expected, as the system must generate sufficient velocity to reach the target pose promptly and accurately.
These results demonstrate that the controller successfully generates the required force profiles to track step-like commands to reach and stay in the desired states.
Despite fluctuations in the estimate of state variables, the actuator forces remain bounded with no signs of instability or oscillatory behavior.
This indicates that the controller is robust to moderate levels of sensor noise without significant degradation in performance.
The consistent and stable force outputs further support the effectiveness of the proposed control scheme, particularly for static reference tracking tasks.

\begin{figure}[!b]
    \centering
        \centering
        \includegraphics[width=.9\linewidth]{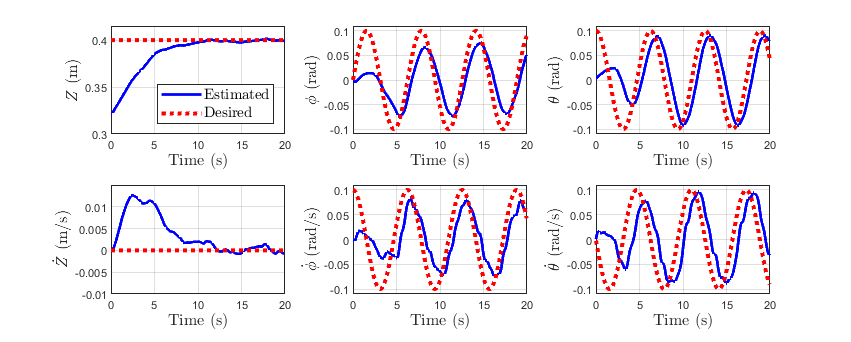}
        \vspace{-5pt}
        \caption{Estimated and desired states of the \textit{sinusoidal motion} scenario.}
       \label{fig:Exp_dancing_pos}
\end{figure}
\begin{figure}[!b]
\vspace{-10pt}
    \centering
    \includegraphics[width=.9\linewidth]{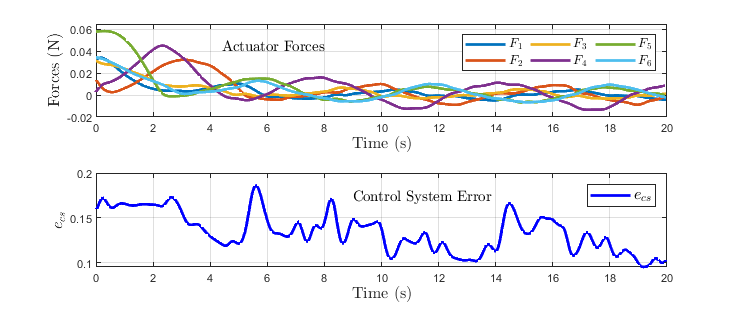}
    \vspace{-5pt}
    \caption{Experimental force control input of the \textit{sinusoidal motion} scenario.}
    \label{fig:Exp_Dancing_Forces}
\end{figure}

\subsubsection{Experimental Result of \textit{Sinusoidal Motion} Scenario}

Fig.~\ref{fig:Exp_dancing_pos} shows the actual and desired state variables of the closed-loop system where the estimated states are plotted as blue lines along with the desired states that are plotted as purple dotted lines.
Only state variables that are subjected to changes in the desired trajectory are shown; the remaining axes are omitted for clarity.
The corresponding force control input and and error observed during the experiment are shown in Fig.~\ref{fig:Exp_Dancing_Forces}.

For the angular motion axes $\phi$ and $\theta$ where dynamic trajectories were applied, the state estimates closely follow the desired values.
The platform's orientation starts from zero and attempts to reach the desired values within specific time steps.
As shown in Fig.~\ref{fig:Exp_dancing_pos}, there is a transient period before the platform finally achieves the target orientation.
Although deviations are present, the overall response is consistent with the desired profile, indicating that the controller effectively handles variations in angular motions.
These deviations can be attributed to practical factors, such as actuator speed constraints, sensor noise, and unmodeled dynamics, which affect the system’s ability to track rapidly changing references.

The estimated velocity results also show that the platform is generally able to follow the desired profile.
For the linear velocity $\dot{Z}$, an overshoot occurs as the platform moves along the $Z$-axis before settling at the target velocity.
This is expected due to the momentum required to reach the desired position.
For the angular velocities $\dot{\phi}$ and $\dot{\theta}$, a gradual transition is observed from the initial to desired velocities, reflecting the response time required by the physical system elements.
Although these fluctuations influence the generated actuator forces, the platform still demonstrates successful tracking of the desired state variables to achieve the required control objective.

\subsection{Limitation}

Although the system generally shows satisfactory performance in tracking both angular and linear trajectories, several practical limitations affect its precision and responsiveness.
State coupling occurs due to actuator mismatches, where the interaction between different DOF influences the overall system dynamics.
Ripple and offset errors are also observed, which can be attributed to actuator backlash, mechanical tolerances, and friction within the drive system; these imperfections cause small repetitive deviations and a persistent bias in the position.
Moreover, during the $Z'$ transient in sinusoidal motion, the platform still exhibits an oscillatory deviation from the desired trajectory in response to rapid changes in reference. 
Furthermore, a phase delay is noticeable between the desired and measured signals, primarily caused by inherent control delays, actuator response time, and the filtering effects of the control algorithm. 
These effects collectively highlight the performance limitations of the real-time platform while still demonstrating satisfactory control performance.


\section{Conclusion}
\label{sec:conclusion}
This paper has presented the complete design, control and experimental validation of a low-cost Stewart platform prototype developed as an affordable yet capable robotic testbed for research and education. 
The platform integrates a feedback linearization scheme with LQR control, supported by EKF-based state estimation, and was validated through both simulation and experimental trials on static and dynamic trajectories.
Simulation and experimental results confirm that the prototype can reliably implement trajectory tracking and state estimation despite sensor noise and disturbances, demonstrating its suitability for research and educational use.
Looking ahead, future work will focus on improving robustness and incorporating safety-critical control strategies to further extend the platform capabilities.


\newpage
\appendix

\section*{Appendix A: BoM and Component Specification of the Stewart Prototype}
\label{sec:apdx}
\begin{table}[!htb]
\small
\caption{The BoM of the Stewart platform prototype. \label{tab:spending}
}
\centering{%
\begin{tabular}{l c l l}
\midrule
\textbf{Components} & \textbf{Quantity} & \textbf{Unit Price} & \textbf{Total} \\
\midrule
Actuonix P-16P  & 6 & \$ 90.00 & \$ 540.00 \\ 
L298N Motor Drivers & 3 & \$ 0.85 & \$  2.56 \\ 
Acrylic (Platform) & 1 & \$ 5.25 & \$ 5.25 \\ 
Acrylic (Base) & 1 & \$ 14.03 & \$ 14.03 \\ 
Universal Joint & 6 & \$ 7.32 & \$ 43.92 \\
Rod End Bearing PHS5 & 6 & \$ 0.64 & \$ 3.84 \\ 
Taubotics ROS IMU & 6 & \$ 125 & \$ 125 \\
NI myRIO-1900 & 1 & \$ 1,156.26 & \$ 1,156.26\\ \midrule
\multicolumn{3}{c}{\textbf{Total}} & \$ 1,890.85 \\
\midrule
\end{tabular}
}%
\vspace{-10pt}
\end{table}

\begin{table}[!htb]
\small
\caption{NI myRIO-1900 hardware specifications.\label{tab:myrio-specs}}
\centering{%
\begin{tabular}{l l}
\midrule
\textbf{Feature} & \textbf{Specifications} \\
\midrule
Power Supply& 6 to 16 VDC \\ 
Processor/FPGA & Xilinx Z-7010 (Dual-core, 667 MHz)\\ 
Memory (Nonvolatile) & 512 MB Flash \\ 
Memory (RAM) & 256 MB DDR3 \\ 
Analog I/O & 6 inputs and 4 outputs with 12-bit resolution \\ 
Digital I/O & 40 channels (3.3 V logic, 5 V tolerant) \\ 
USB Ports & USB 2.0 Hi-Speed \\
\midrule
\end{tabular}
}%
\end{table}


\section*{Appendix B: Derivation of Dynamics Model}

\subsection{Additional Kinematic Model Derivations}
The unit vector of each Stewart platform leg can be calculated through:
\begin{equation}
    n_i=\frac{l_i}{s_i}.
\end{equation}
Additionally, the position vector of the joint platform w.r.t frame \{B\} is computed by the following:
\begin{equation}
    q_p = t+Rp_i
\end{equation}
and the time derivative is given by the following:
\begin{equation}
	\dot{q}_{pi}=\begin{bmatrix}
		\mathrm{I} & {R\tilde{p_i}^\intercal R^\intercal}
	\end{bmatrix}
	\dot{q}.
	\label{eqn:qp_dot}
\end{equation}
Based on \eqref{eqn:qp_dot}, the velocity of the legs is given by:
\begin{equation}
	\dot{s}_i = n_i^\intercal\dot{q}_{pi}.
	\label{eqn:leg_speed}
\end{equation}
On the other hand, the velocities $v_{ti}$ and $v_{bi}$ of the upper and lower parts of the Stewart platform legs, respectively, can be computed as follows.
\begin{equation}
 \label{eqn:vb}
    v_{ti}= \left(\mathrm{I}+\frac{l_t \tilde{n_i}^2}{s_i}\right)\dot{q}_{pi}, \qquad
    v_{bi}= \left(\frac{l_b \tilde{n_i}^\intercal\tilde{n_i}}{s_i}\right)\dot{q}_{pi}.
\end{equation}
Finally, the angular velocity of the Stewart platform leg is defined by the following.
\begin{equation}
    \omega_{li} = \frac{\tilde{n}_i \dot{q}_{pi}}{s_i}.
	\label{eqn:wl} 
\end{equation}

\subsection{Legs Dynamic Model Derivation}
For this subsection, the symbol $i$ which is used to denote the leg number will be omitted to avoid confusion.
The Lagrange equation of the Stewart platform leg is defined as follows:
\noindent\begin{equation}
	\frac{d}{dt}\left(\frac{\partial{T}}{\partial{\dot{q}_p}}\right) - \frac{\partial{T}}{\partial{q_p}} = Q,
	\label{eqn:lagrange_T}
\end{equation}

\noindent where $T$ is the kinetic energy which is defined by:
\noindent\begin{align}
	T &=  \frac{1}{2} v_{t}^\intercal m_{t} v_{t} + \frac{1}{2} \omega_l^\intercal\left(\mathrm{I}_{t}+\mathrm{I}_{b}\right) \omega_l \nonumber \\
	&=  \frac{1}{2} \dot{q}_p^\intercal \left(\left(\mathrm{I}+\frac{l_{\mathrm{t}} \tilde{n}^2}{s}\right)^\intercal m_t \left(\mathrm{I}+\frac{l_{\mathrm{t}} \tilde{n}^2}{s}\right) +\left(\mathrm{I}_t+\mathrm{I}_b \right) \tilde{n}^\intercal \frac{\tilde{n}}{s^2}\right) \dot{q}_p \nonumber \\
	&=  \frac{1}{2} \dot{q}_p^\intercal\left(\mathrm{M}_1 + \mathrm{M}_2\right) \dot{q}_{\mathrm{p}}
	\label{eqn:T}.
\end{align}

\noindent Here, $\mathrm{M}_1$ and $\mathrm{M}_2$ are inertial term defined as follows:
\noindent\begin{equation}
    \mathrm{M}_1 = \left(\mathrm{I}+\frac{l_{\mathrm{t}} \tilde{n}^2}{s}\right)^\intercal m_t \left(\mathrm{I}+\frac{l_{\mathrm{t}} \tilde{n}^2}{s}\right) ,\qquad
    \mathrm{M}_2 = \left(\mathrm{I}_t+\mathrm{I}_b \right) \tilde{n}^\intercal \frac{\tilde{n}}{s^2}.
\end{equation}

Next, we define the forces acting upon the leg as follows.
\begin{equation}
	Q_f = nf,\qquad
	Q_{m_tg} = \left(\mathrm{I} + \frac{l_t \tilde{n}^2}{s} \right) m_t g,\qquad
	Q_{m_bg} = \left(\frac{l_b \tilde{n}^{\mathrm{T}}n}{s} \right) m_b g.
	\label{eqn:Qmbg}
\end{equation}
The generalized force is defined by:
\begin{equation}
	Q = Q_f + Q_{m_tg} + Q_{m_bg} + f_p.
	\label{eqn:Q}
\end{equation}
The constraint force $f_p$ can be rewritten into \eqref{eqn:fp_i} with the function $\mathrm{C}_a$ below.
\begin{equation}
    \noindent\begin{aligned}
        \mathrm{C}_a = &\frac{m_t l_t}{s^2}(n \dot{q}_p^\intercal  \tilde{n}^\intercal  n + n^\intercal  \dot{q}_p \tilde{n}^\intercal  \tilde{n} + \tilde{n}^\intercal \tilde{n}\dot{q}_p n^\intercal  ) 
        - \frac{m_t l_t^2}{s^3}(n^\intercal  \dot{q}_p \tilde{n}^\intercal  \tilde{n} + \tilde{n}^\intercal \tilde{n}\dot{q}_p n^\intercal  )
        - \frac{2(\mathrm{I_t} + \mathrm{I_b})}{s^3}(\tilde{n}^\intercal \tilde{n}\dot{q}_p n^\intercal  ).
    \end{aligned}
    \label{eqn:C_a}
\end{equation}

\subsection{Platform Dynamic Model Derivation}
Combining equations \eqref{eqn:platform_newton} and \eqref{eqn:platform_newton_base} will produce the following equation:
\noindent\begin{equation}
	\mathrm{M}_p \ddot{q} + \mathrm{C}_p \dot{q} + \noindent\begin{bmatrix}
		m_p \mathrm{I} \\
		m_p R \tilde{c_p}^\intercal R^\intercal
	\end{bmatrix} \tilde{\omega}^2 R c_p = \mathrm{H}_p \mathrm{F}_p + \noindent\begin{bmatrix}
		m_p \mathrm{I} \\
		m_p R \tilde{c_p}^\intercal R^\intercal g
	\end{bmatrix},
	\label{eqn:platform_dyn}
\end{equation}
\noindent where the matrices and vectors in the model are given as follows.
\noindent\begin{align}
    \mathrm{M}_p& = \noindent\begin{bmatrix}
        m_p \mathrm{I} & m_p R\tilde{c}_p^\intercal  R^\intercal  \\
        m_p R \tilde{c}_p R^\intercal  & m_p R \tilde{c}_p \tilde{c}_p^\intercal  R^\intercal  + R \mathrm{I}_p R^\intercal 
    \end{bmatrix},
    \label{eqn:Mp}\\
    \mathrm{C}_p &= \noindent\begin{bmatrix}
        0 & 0 \\
        0 & \tilde{\omega}R\mathrm{I}_pR^\intercal 
    \end{bmatrix},\\
    \mathrm{H}_p &= \noindent\begin{bmatrix}
        \mathrm{I} & \dots & \mathrm{I} \\
        (R\tilde{p}R^\intercal )_1 & \dots &(R\tilde{p}R^\intercal )_6
    \end{bmatrix},
    \label{eqn:H_p}
    \\
    \mathrm{F}_p &= \noindent\begin{bmatrix}
        f_{p1} & f_{p2} & f_{p3} & f_{p4} & f_{p5} & f_{p6}
    \end{bmatrix}^\intercal .
\end{align}
The complete dynamic equation that combines the platform and legs models is defined in \eqref{eqn:stewart_dyn_equation}, in which the details of each matrices are defined as follows.
\begin{align*}
    \mathrm{M}(q) &= \mathrm{M}_p + \sum_{i=1}^6 
    \begin{bmatrix}
        \mathrm{I} \\
        R \tilde{p}_i R^\intercal
    \end{bmatrix}
    (\mathrm{M}_1 + \mathrm{M}_2)_i
    \begin{bmatrix}
        \mathrm{I} & R \tilde{p}_i R^\intercal
    \end{bmatrix},
    \nonumber \\[1ex]
    \mathrm{C}(q,\dot{q})\dot{q} &= \mathrm{C}_p \dot{q} + \sum_{i=1}^6 
    \begin{bmatrix}
        \mathrm{I} \\
        R \tilde{p}_i R^\intercal
    \end{bmatrix}
    (\mathrm{C}_a)_i
    \begin{bmatrix}
        \mathrm{I} & R \tilde{p}_i R^\intercal
    \end{bmatrix}
    \dot{q}  + 
    \begin{bmatrix}
        m_p \mathrm{I} \\
        m_p R \tilde{c}_p^\intercal R^\intercal
    \end{bmatrix}
    \tilde{\omega}^2 R c_p 
    + \sum_{i=1}^6 
    \begin{bmatrix}
        \mathrm{I} \\
        R \tilde{p}_i R^\intercal
    \end{bmatrix}
    (\mathrm{M}_1 + \mathrm{M}_2)_i \tilde{\omega}^2 (R p_i),\\[1ex]
    \mathrm{G}(q) &= - 
    \begin{bmatrix}
        m_p \mathrm{I} \\
        m_p R \tilde{c}_p^\intercal R^\intercal g
    \end{bmatrix}
     - \sum_{i=1}^6 
    \begin{bmatrix}
        \mathrm{I} \\
        R \tilde{p}_i R^\intercal
    \end{bmatrix}
    (Q_{m_{tg}} + Q_{m_{bg}})_i,\\[1ex]
    \mathrm{H}(q) &= 
    \begin{bmatrix}
        n_1 & \dots & n_6 \\
        (R \tilde{p}_1 R^\intercal) n_1 & \dots & (R \tilde{p}_6 R^\intercal) n_6
    \end{bmatrix},\\[1ex]
    \mathrm{F} &= 
    \begin{bmatrix}
        f_1 & f_2 & f_3 & f_4 & f_5 & f_6 
    \end{bmatrix}^\intercal.
\end{align*}

\bibliographystyle{IEEEtran}
\bibliography{stewart2025}

\end{document}